\pdfoutput=1

\documentclass[11pt]{article}

\usepackage[]{acl}

\usepackage{times}
\usepackage{latexsym}

\usepackage[T1]{fontenc}

\usepackage[utf8]{inputenc}

\usepackage{microtype}

\title{A Kind Introduction to Lexical and Grammatical Aspect,\\ with a Survey of Computational Approaches}

\usepackage{xspace}
\usepackage{booktabs}
\usepackage{graphicx}
\usepackage{enumitem}
\usepackage{xcolor,colortbl}
\usepackage{tikz}
\usepackage{tikz-qtree}
\usepackage{tikz-dependency}
\usepackage{url}
\usepackage{amssymb}
\usepackage{float}
\usepackage{todonotes}
\usepackage{csquotes}

\usepackage{multirow}
\usepackage{tabularx}
    \newcolumntype{L}{>{\raggedright\arraybackslash}X}

\usepackage{caption}
\usepackage{subcaption}

\usepackage{glossaries}

\setlist[description]{leftmargin=\parindent,labelindent=0pt}

\newcommand{\sref}[1]{Section~\ref{sec:#1}}
\newcommand{\aref}[1]{Appendix~\ref{sec:#1}}

\newcommand{\tref}[1]{Table~\ref{tab:#1}}
\newcommand{\fref}[1]{Figure~\ref{fig:#1}}
\newcommand{\eref}[1]{(\ref{ex:#1})}
\newcommand{\eeref}[2]{(\ref{ex:#1}#2)}

\newcommand{\dul}[1]{\underline{\smash{#1}}}

\definecolor{anne}{rgb}{0.635,0.998,0.722}
\definecolor{alexis}{rgb}{0.06, 0.75, 0.99}
\definecolor{nianwen}{rgb}{1, 0.8, 0.1}

\definecolor{mdgreen}{rgb}{0,0.6,0}
\definecolor{orange}{rgb}{1,0.5,0}

\newcounter{example}
\newenvironment{example}[1][]{\refstepcounter{example}\par\hspace*{-3mm}
	\textbf{(\theexample#1)} \rmfamily}{}

\usepackage{array}
\newcolumntype{C}[1]{>{\centering\let\newline\\\arraybackslash\hspace{0pt}}m{#1}}
\newcolumntype{R}[1]{>{\raggedleft\let\newline\\\arraybackslash\hspace{0pt}}m{#1}}

\usepackage{pinyin}

\newcommand{\comm}[1]{[\textcolor{blue}{\textit{#1}}]}

\author{Annemarie Friedrich$^1$ \hspace{5mm}
  Nianwen Xue$^2$
  \hspace{5mm}
  Alexis Palmer$^3$ \\
  $^1$Bosch Center for Artificial Intelligence, Renningen, Germany\\
  $^2$Brandeis University \hspace{4mm}
  $^3$Dept. of Linguistics, 
  University of Colorado Boulder\\
\texttt{annemarie.friedrich@de.bosch.com} \hspace{4mm}  \texttt{xuen@brandeis.edu} \\
\texttt{alexis.palmer@colorado.edu}
}  

\begin{document}

\maketitle

\begin{abstract}
Aspectual meaning refers to how the internal temporal structure of situations is presented.
This includes whether a situation is described as a state or as an event, whether the situation is finished or ongoing, and whether it is viewed as a whole or with a focus on a particular phase.
This survey gives an overview of computational approaches to modeling lexical and grammatical aspect along with intuitive explanations of the necessary linguistic concepts and terminology.
In particular, we describe the concepts of stativity, telicity, habituality, perfective and imperfective, as well as influential inventories of eventuality and situation types.
Aspect is a crucial component of semantics, especially for precise reporting of the temporal structure of situations, and future NLP approaches need to be able to handle and evaluate it systematically.
\end{abstract}

\section{Introduction}
\label{sec:intro}

Lexical and grammatical aspect play essential roles in semantic interpretation \citep{smith2003modes}, and yet even state-of-the-art natural language understanding (NLU) systems do not address these linguistic phenomena systematically \citep{metheniti-etal-2022-about}.
Consider this example: an NLU-based personal assistant, noticing the boarding time of a flight, tells a passenger (who is still shopping at the airport) \enquote{You miss flights} (i.e. on a regular basis) instead of \enquote{You are missing the flight} (now!).
The traveler might misinterpret this utterance as chit-chat and indeed miss the flight.
Aspectual encoding errors impair fluid and correct communication.

While there has been a notable amount of work on modeling lexical and grammatical aspect in the computational linguistics community \cite{moens-steedman-1988-temporal,siegel-mckeown-2000-learning,friedrich-etal-2016-situation,kober-etal-2020-aspectuality}, this area is still a niche in natural language processing (NLP).
In this paper, we survey the state of research in this area and argue that a good computational understanding of lexical and grammatical aspect is paramount for capturing temporal information signaled by linguistic encoding.
We thereby add to the on-going discussion in the NLP community about what is required to achieve true natural language understanding \cite{bender-koller-2020-climbing,dunietz-etal-2020-test,trott-etal-2020-construing}.

\begin{figure}
    \centering
    \footnotesize
    \scalebox{0.9}{
    \begin{tikzpicture}
    \node (g1) at (-1.1,0) {\includegraphics[width=2cm]{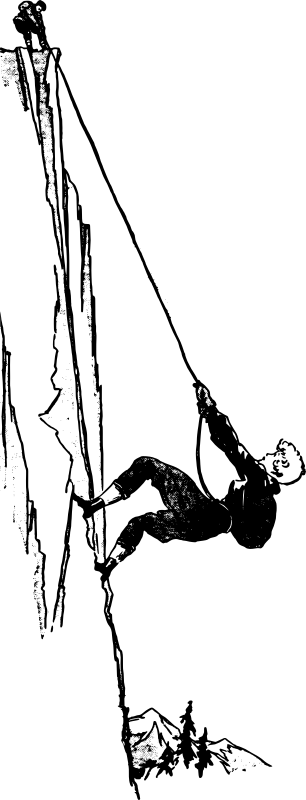}};
    \draw[thick,dashed] (0,-1.2) -- (0,0.2);
    \draw[thick,dashed] (0,-.5) -- (1,-.5);
    \node (t1) at (2,-0.5) {\textit{Tom is climbing.}};
    \node (t11) at (3,-0.8) {ongoing event, progressive, \textbf{activity}};
    \draw[thick] (0.2,-2.7) -- (0.2,2.3);
    \draw[] (0.2,.8) -- (0.5,.8);
    \node (t2) at (3,0.8) {\textit{Tom climbed the mountain (yesterday).}};
    \node (t22) at (3.1,0.5) {dynamic, completed event, perfective};
    \node (t24) at (3.1,0.2) {telic = has endpoints, \textbf{accomplishment}};
    \node (t3) at (2,2.5) {\textit{John is standing/stood on the cliff.}};
    \draw[thick,densely dashdotted] (6,-2.7) -- (6,2.7);
    \draw[thick,densely dashdotted] (-2,2.5) -- (-.2,2.5);
    \draw[thick,densely dashdotted] (4.3,2.5) -- (6,2.5);
    \node (t31) at (3,2.2) {no change, \textbf{state}};
    \node (t4) at (3,1.7) {\textit{At 4:07pm, Tom reached the top.}};
    \node (t41) at (3,1.4) {punctual, change of state, \textbf{achievement}};
    \draw[-latex] (0.9,1.7) -- (-1.7,2.3);
    \draw [draw=black] (0.5,-2.8) rectangle (5.5,-1.5);
    \node (i1) at (1,-2) {\includegraphics[width=7mm]{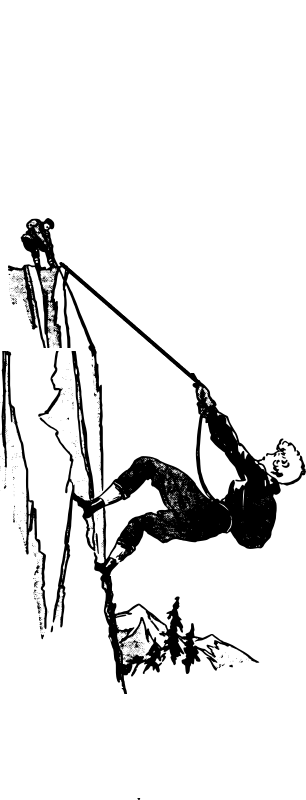}};
    \node (i1) at (2,-2) {\includegraphics[width=7mm]{img/climbingsmaller.png}};
    \node (i1) at (3,-2) {\includegraphics[width=7mm]{img/climbingsmaller.png}};
    \node (t5) at (4.5,-1.8) {\textit{Tom climbs}};
    \node (t51) at (4.5,-2.1) {\textit{mountains.}};
    \node (t51) at (4.5,-2.4) {\textbf{habitual}};
    \end{tikzpicture}}
    \caption{\textbf{Aspect} is like the camera lens of language, the device by which we focus on particular phases of a situation \citep{vendler1957verbs,smith1997parameter}.}
    \label{fig:teaser}
\end{figure}

\begin{table*}[t]
    \centering
    \footnotesize
    \setlength\tabcolsep{3pt}
\begin{tabular}{lllll}
\toprule
\textbf{Dataset} & \textbf{Reference} & \textbf{Language} & \textbf{Size} & \textbf{Annotated Categories}\\
\midrule
    \href{https://www.coli.uni-saarland.de/projects/sitent}{Asp-Ambig} & \citet{friedrich-palmer-2014-automatic} & EN & 7875 clauses & stativity\\
    \href{https://github.com/annefried/sitent}{SitEnt} & \citet{friedrich-etal-2016-situation} & EN & 40,000 clauses & situation entities, stativity, habituality \\
    \href{https://www.idiap.ch/en/dataset/tense-annotation}{Tense-Europarl} & \citet{loaiciga-grisot-2016-predicting} & EN/FR & 435 verb phrases & boundedness (in sentence context)\\
    \href{https://github.com/annefried/telicity}{MASC-telicity} & \citet{friedrich-gateva-2017-classification} & EN & 1863 clauses & telicity \\
    \href{https://github.com/malihealikhani/Captions}{Captions} & \citet{alikhani-stone-2019-caption} & EN & 2600 captions & stative, durative, telic/atelic, punctual\\
    \href{https://github.com/wroberts/annotator}{SdeWac-Aspect} & \citet{egg-etal-2019-annotation} & DE & 4200 clauses & stativity, durative vs. punctual, boundedness\\ 
    \href{ https://go.rutgers.
edu/cb6le5c1.}{DIASPORA} & \citet{kober-etal-2020-aspectuality} & EN & 927 utterances & stative, telic, atelic \\
    \href{http://decomp.io/data/}{UDS} & \citet{gantt2022decomposing} & EN & 16,624 sentences; & stativity, some telicity, durativity, \\
    & & & granularity varies & other event properties \\
\bottomrule
\end{tabular}
    \caption{Available \textbf{Datasets} labeled with aspectual information.}
    \label{tab:aspect-datasets}
\end{table*}

When describing a situation, as exemplified in \fref{teaser}, producers of language have options for how to depict the situation and which subparts of it to highlight.
Just as a cinematographer uses focus to highlight certain elements of a scene and thus bring them to the attention of the viewer, different aspectual choices focus on different subparts of the situation at hand, leaving the rest as background \cite[analogy due to][]{smith2003modes}.
For example, many situations can be construed as either \textit{foregrounded} events (\enquote{Tom climbed the mountain}), moving narrative time forward, or as states (\enquote{John stood on the cliff}), which often function as the \textit{background} in a narrative or explanation.
Aspectual choices may trigger hard inferences, softer implicatures, or simply expectations, having a non-negligible impact on the reader or listener.
One aim of this survey is to clarify the sometimes complex linguistic work on aspect, giving NLP researchers and practitioners a conceptual toolkit, plus a glossary of aspect-related terminology (Appendix~\ref{sec:glossary}).
The second aim is to survey computational work on modeling aspect.

Aspectual categories are \textit{semantic} notions that are conveyed through a variety of mechanisms (\sref{aspect-live}).
Fundamental aspectual distinctions (\sref{distinctions}) may be part of the lexical meaning of a verb and are also influenced by context. 
These distinctions drive some influential inventories of eventuality types (\sref{vendler}).
Grammatical aspect (\sref{grammAspect}) is marked morphologically in some (e.g., Slavic) languages, or via the choice or absence of tense-aspect features in languages such as English.
\sref{sitent} explains the idea of situation types, aspectual distinctions at the clause level.

As a consequence of the often distributed or even implicit encoding of aspect, computational models need to consider a variety of lexical and grammatical features as well as discourse and pragmatic information.
While some NLP systems trained on end-to-end datasets may get some of these cases right, we are not aware of a system treating such problems in a principled manner.
A recent study finds that transformer-based embeddings are useful for classifying clausal aspect \citep{metheniti-etal-2022-about}, but it remains unclear just how these models learn or incorporate aspectual information.
This survey focuses on modeling aspectual information as it is annotated in the datasets listed in \tref{aspect-datasets}.

In \sref{analysis}, we discuss potential ways of moving the field forward.
We argue that explicit modeling or evaluation of aspectual categories should become a routine  perspective in future NLP: getting aspect right has high stakes for applications such as temporal question answering, machine translation, or computer-aided language learning.

\section{Where Does Aspect Live?}
\label{sec:aspect-live}

\citet{comrie1976aspect} notes that
there is a terminological and conceptual confusion around tense and aspect.
This is at least partially because the expression of temporal location is intertwined morphologically with aspect in many languages of the world \citep{smith1997parameter}, so the traditional grammatical terminologies of individual languages often introduce tense/aspect combinations simply as tenses.
Tense is \textit{deictic} and locates the time of a situation relative to the utterance or relative to another situation.
Aspect, in contrast, refers to situation-internal time.

Aspectual distinctions are \textit{covert} categories \citep{whorf1945grammatical,smith1997parameter}, expressing particular temporal semantic concepts in a way that does not have a one-to-one mapping to lexical, grammatical or syntactic categories \citep{fillmore1969types}. %
For example, the English Progressive\footnote{Following \citet{comrie1976aspect}, we use initial capitals for the names of language-particular categories and lower case for language-independent semantic distinctions.} often indicates an ongoing situation without focusing on its endpoints (\enquote{I am eating an apple}), but it also 
signals future events (\enquote{I am going to Paris}).
The grammatical form of such tense-aspect combinations can often be automatically detected with high accuracy \citep{ramm-etal-2017-annotating,myers-palmer-2019-cleartac}.
This type of aspect, encoded in the tense system of a language, is different from \textit{semantic} concepts which refer to how the internal structure of a situation is presented; this survey focuses on the latter.

\textbf{So where does \textit{covert, semantic} aspect live?}
Aspectual meaning 
is compositional, composed by the verb’s inherent meaning, its arguments, its tense, any morphological aspect markers, some prepositional phrases, and the adverbs of the sentence \citep{verkuyl1972compositional,mourelatos1978events,smith1997parameter}.
Eventuality and situation type inventories make use of several dimensions of aspect to distinguish between different types of temporal structures.
\textit{Lexical aspect} refers to 
information contained in the meaning of verbs (or verb senses),
with the consequence that their values often can be determined only when observing how the verb interacts with its clausal  context. %
A second dimension, 
\textit{grammatical} or \textit{viewpoint} aspect, refers to whether there is a focus on part of the situation or whether the situation is viewed as a whole or even as repeating.
Grammatical aspect and situation types are both semantic categorizations at the sentence/clause level \citep[e.g.,][]{vendler1957verbs,verkuyl2005aspectual,smith2003modes}.

\section{Inherent Lexical Aspect}
\label{sec:distinctions}

Here we explain two fundamental %
notions related to inherent \textit{lexical aspect}: \textit{stativity} refers to whether a situation is viewed as an event or state, and \textit{telicity} refers to whether an endpoint is visible in a clause.

\subsection{Stativity}\label{sec:stativity}
The most fundamental distinction made in the hierarchies of eventuality types is that between \textit{states} (\enquote{love,} \enquote{own}) and \textit{events} (\enquote{run,} \enquote{buy}).
In contrast to \textit{dynamic} predicates, \textit{stative} verbs entail no change \citep{filip2012lexical}.
States obtain in time but they do not take time; events occur, happen, or take place \citep{smith1997parameter}.
Events may have a beginning and/or an end \citep{comrie1976aspect}, and they are often \textit{durative}, i.e., conceived as lasting for a certain period of time.
Some events are dynamic situations describing a change of state such as \enquote{John \dul{reached} the goal}; these are perceived as \textit{punctual}. %
For punctual, single-stage events which cause no change of state, \citet{smith1997parameter} adds the situation type \textit{semelfactives}, for verbs like \enquote{knock,} \enquote{flash,} or \enquote{blink.}
While dynamic in nature, semelfactives return to their initial state at their end and often occur as \textit{iteratives} \citep{filip2012lexical}.
The stativity distinction pertains to verb senses, not verb types.
For example, the verb type \enquote{make} has both a dynamic \eeref{make}{a} and a stative \eeref{make}{b} sense.

\begin{example}
(a) She is \textbf{making} a cake. (\textit{dynamic})\\
\hspace*{7.3mm}(b) She \textbf{makes} a great host. (\textit{stative})
\label{ex:make}
\end{example}

In theory, the analysis of stativity could be applied to inventories of verb senses such as WordNet \citep{fellbaum2010wordnet}.
In practice, most computational works focus on classifying verbs in context.

\textbf{Early computational approaches.}
The dynamic-stative distinction is at the heart of early approaches to computational modeling of tense and aspect.
In the PUNDIT system for temporal information processing \citep{passonneau-1988-computational}, the lexical aspect of verbs in tensed clauses can be read from the system's output: transition events use \texttt{become}, processes use \texttt{do}; other cases signal states.
In the context of lexicon induction, \citet{klavans-chodorow-1992-degrees} suggest representing event structure of a verb as its \textit{degree of stativity}, which is estimated by the proportion of occurrences in a corpus that are in the Progressive.
\citet{brent-1991-automatic} also presents a program for identifying stativity using syntactic indicators: verbs that occur frequently with the Progressive or that combine with rate adverbs (e.g. \enquote{quickly}, \enquote{slowly}) are usually dynamic.
Both works evaluate by manually inspecting system output. %

\textbf{Datasets.}
\citet{siegel-1999-corpus} presents a manually annotated dataset of 739 training and 739 test clauses taken from medical discharge summaries, covering 222 different verb types. %
With the aim of reproducing and extending their work, \citet{friedrich-palmer-2014-automatic} have three annotators label 7,875 clauses from MASC \citep{ide-etal-2008-masc}, marking the clause's main verb as \textit{stative} or \textit{dynamic}.%
If annotators see both readings, they may select both.
The dataset has later been extended to cover 30,333 clauses from MASC and 10,607 clauses from Wikipedia \citep{friedrich-etal-2016-situation}.
\citet{kober-etal-2020-aspectuality} introduce the DIASPORA dataset, in which 927 utterances from a corpus of human-human phone conversations \citep{brennan2013entrainment} are labeled with whether the first verb phrase of each utterance is stative or dynamic.
The dataset spans 69 different verb types.
\citet{chen-palmer-2022-contrast} produce 292 contrast sets for stativity in English and outline strategies for converting between stative and dynamic expressions of the same situation.
For more datasets that annotate stativity as one feature among others, see \sref{vendler}.

\textbf{Modeling.}\footnote{Appendix~\ref{sec:overview-modeling} summarizes computational systems and approaches to modeling aspect.}
\citet{siegel-1999-corpus,siegel-1997-learning} and \citet{siegel-mckeown-2000-learning} propose a machine-learning approach to classifying stativity that describes each verb occurrence exclusively using 
corpus-based statistics of the corresponding verb type.
The verb-type based indicators are normalized counts that reflect, for instance, how often the verb co-occurs with the past tense, in the perfect, or in negated form
(for a full list, see \aref{ling_ind}).
Using these features, they train logistic regression models, decision trees, and genetic programming algorithms.

\citet{friedrich-palmer-2014-automatic} compare these corpus-based \textit{linguistic indicators} to instance-based syntactic-semantic features representing the clausal context, e.g., the part-of-speech tag of the verb, tense, voice, and WordNet information \citep{fellbaum2010wordnet} for the verb's arguments.
Using the LCS database \citep{dorr2001lcs} and a procedure described by \citet{dorr-olsen-1997-deriving}, they also construct three seed sets with verb types that either occur exclusively as \textit{dynamic}, only as \textit{stative}, or can take both aspects.
Based on these seed sets and a pre-trained syntactic distributional model \citep{thater-etal-2010-contextualizing}, similarity values are computed and used as additional features.
\citet{kober-etal-2020-aspectuality} input sums of non-contextualized word embeddings \citep{mikolov2013efficient} for the clause's main verb as well as selected context words to a logistic regression classifier.
Overall, they find that using the local context in the form of a word window outperforms the verb-type only classifier, but that feeding in the entire sentence hurts.
\citet{metheniti-etal-2022-about} also classify stativity
(called \enquote{duration} in their work)
by fine-tuning various transformer models and classification layers.
They find BERT \citep{devlin-etal-2019-bert} models work better than RoBERTA \citep{roberta}, AlBERT \citep{albert} and XLNet \citep{yang2019xlnet}.

\subsection{Telicity}
The second important distinction related to inherent lexical aspect is that of \textit{telicity}.
The term \textit{telic}, derived from Greek \textit{télos} (goal) was introduced by \citet{garey1957verbal}.
In his definition, telic verb senses have a built-in goal: when that goal is reached, a change of state occurs and the event is complete \citep{smith1997parameter}.
Telicity is also sometimes referred to as \textit{boundedness} \citep[e.g., by][]{loaiciga-grisot-2016-predicting}.
As illustrated by \eref{lake-swim}, telicity is a feature of the entire clause \citep{verkuyl2005aspectual}.

\begin{example}
(a) He was swimming in the lake. \textit{(atelic)}\\
\hspace*{7.5mm}(b) He was swimming across the lake. \textit{(telic)}
\label{ex:lake-swim}
\end{example}
When a telic verb is used in the \textit{imperfective} as in \eeref{telic-recover}{a}, the arrival or nonarrival at the goal is hidden \citep{garey1957verbal}.
If the same verb is applied in the \textit{perfective} as in \eeref{telic-recover}{b}, it means that the goal has been reached at the time of reference.
Hence, if \eeref{telic-recover}{a} is true at some particular point in time, it cannot be the case that \eeref{telic-recover}{b} is true at the same point in time.

\begin{example}
(a) John was recovering. \textit{(telic, imperfective})\\
\hspace*{7.5mm}(b) John has recovered. \textit{(telic, perfective)}
\label{ex:telic-recover}
\end{example}

In contrast, \textit{atelic} verbs do not have to wait for a goal for their realization; they are realized as soon as they begin.
If an atelic verb is used in imperfective form as in \eeref{atelic-sing}{a}, we can infer that the sentence in perfective form \eeref{atelic-sing}{b} is also true.

\begin{example}
(a) Sue was singing. \textit{(atelic, imperfective})\\
\hspace*{7.5mm}(b) Sue has sung. \textit{(atelic, perfective)}
\label{ex:atelic-sing}
\end{example}

\textbf{Datasets.}
\citet{siegel-mckeown-2000-learning} describe a small dataset annotated for \textit{completedness}, i.e, telicity.
Their training and test sets each consist of approximately 300 clauses taken from 10 novels and covering 204 different dynamic verbs. %
\citet{friedrich-gateva-2017-classification} manually annotate 1863 clauses taken from MASC \citep{ide-etal-2008-masc} for telicity.
The Captions dataset \citep{alikhani-stone-2019-caption} marks telicity in several image caption corpora.
They find that the proportion of telic verbs ranges from 6\% to 59\% across corpora, with atelic descriptions prevailing in almost all corpora.
In the DIASPORA dataset \citep{kober-etal-2020-aspectuality}, in each utterance, the first VP is annotated with its \textit{predicational} aspect (stative, telic, or atelic).

\textbf{Modeling.}
\citet{siegel1996gathering,siegel-mckeown-2000-learning} propose a machine-learning approach to classifying telicity of verb occurrences
using their above-described corpus-based linguistic indicators.
For example, verbs frequently occurring in the Progressive are likely atelic.
Leveraging these corpus-based \textit{linguistic indicators}, \citet{friedrich-gateva-2017-classification}  integrate syntactic-semantic contextual features into their logistic regression model.
They also leverage additional silver standard training data in the form of projected annotations from the English-Czech InterCorp \citep{ek2012case,rosen-vavrin-2012-building}.
Their approach, using the machine-readable Czech dictionary Vallex \citep{vzabokrtsky2007valency}, is based on the assumption that most perfective Czech verbs are translated using telic verb constructions, and that most imperfective verbs correspond to atelic constructions. %
\citet{loaiciga-grisot-2016-predicting} create an automatic classifier similar to that of \citet{friedrich-gateva-2017-classification} to classify \textit{boundedness} of French verbs, i.e., whether the endpoint of an event has occurred or not.
They show that this is useful for picking the correct tense in French translations of the English Simple Past.
Several more recent studies have shown that distributional and neural models can be trained to predict telicity as annotated in available datasets \citep{kober-etal-2020-aspectuality,metheniti-etal-2021-predire,metheniti-etal-2022-about}.
BERT-style models perform well on existing telicity datasets (with larger models outperforming smaller models), yet it is still unclear how or whether they actually capture aspect.
\citet{metheniti-etal-2022-about} observe that models are always highly confident in their predictions, regardless of their accuracy (a common phenomenon for neural models).
We are not aware of any probing or explainability studies in this area yet, and how to obtain better calibrated models is an open research question.

\subsection{Eventuality Types}
\label{sec:vendler}

\begin{table}[t]
\centering
\footnotesize
\setlength\tabcolsep{3pt}
\begin{tabular}{l|ccc|l}
\toprule
 \textbf{Type} & \rotatebox[origin=c]{90}{\parbox[c]{1cm}{\textit{dynamic}}} & \rotatebox[origin=c]{90}{\parbox[c]{1cm}{\textit{durative}}} & \rotatebox[origin=c]{90}{\parbox[c]{1cm}{\textit{telic}}} & \textbf{Examples}\\
\midrule
state & - & + & - & know the answer, love Mary\\
activity & + & + & - & laugh, stroll in the park, swim\\
accomplishment & + & + & + & build a house, walk to school\\
achievement & + & - & + & win a race, recognize\\
semelfactive & + & - & - & tap, knock, hiccup, tap, wink\\
\bottomrule
\end{tabular}
\caption{Eventuality types \citep{vendler1957verbs,smith1997parameter}.}
\label{tab:vendlerClasses}
\end{table}

We now explain some influential inventories of \textit{eventuality types}, which are ontologies of the temporal structures of events and/or states.
\tref{vendlerClasses} shows the set of situation types introduced by \citet{vendler1957verbs}: \textit{state}, \textit{activity}, \textit{accomplishment}, and \textit{achievement}.
There are other similar schemata \citep[][more details in \aref{eventtypes}]{kenny1963action,mourelatos1978events}; we use Vendler's terminology here.

\textit{States} are inherently stative and atelic.
They are \textit{durative} as they usually extend in time, though this time period may be very short as in \eref{shortState}.

\begin{example}
He \dul{was} very \dul{quiet} for two seconds.
\label{ex:shortState}
\end{example}

The other three Vendler classes are all dynamic.
They differ in whether they have a built-in endpoint, and in whether 
a clearly defined process leads up to this endpoint.
\textit{Activities} as in \eref{activity}, consisting entirely of a process, use atelic verbs.
\textit{Accomplishments} as in \eref{accomplishment} consist of a process that leads up to a built-in terminal point.
Similarly, \textit{achievements} as in \eref{achievement} have an endpoint including a change of state, but the verb meaning does not include a process leading up to this point.
\citet{smith1997parameter} adds the \textit{semelfactive type} (see \sref{stativity}).

\begin{example}
Mary was laughing. \textit{(activity)}
\label{ex:activity}
\end{example}
\begin{example}
Mary wrote a letter. \textit{(accomplishment)}
\label{ex:accomplishment}
\end{example}
\begin{example}
He arrived at the station. \textit{(achievement)}
\label{ex:achievement}
\end{example}

\citet{moens-steedman-1988-temporal} work with aspectual profiles of sentences (for details see \aref{eventtypes}) that are classified by making reference to a so-called \textit{nucleus}. %
For example, \enquote{John built a house} consists of the \textit{preparatory phase} (the house is being built), a \textit{culmination point} (the moment at which it is completed), and a \textit{consequent state} (the house is complete).
The process of \textbf{aspectual coercion} shifts aspectual verb types based on their arguments or other aspectual operators such as adverbials.
When used with a predicate whose lexical entry corresponds to a culminated process, the English Progressive strips off the culmination point and makes visible only the preparatory process as in \enquote{John was running a mile.}\footnote{Thus offering an elegant solution to the \textit{imperfective paradox} \citep{dowty1979word,lascarides1991progressive}.}

\textbf{Datasets.}
\citet{zarcone-lenci-2008-computational} create a dataset of 3129 occurrences of 28 Italian verbs manually annotated for Vendler-style event types.
\citet{richard2012lexical} works on an eight-way classification task for categories based on \citeauthor{leech1971meaning}'s \citeyearpar{leech1971meaning} classes (see \aref{leech}). %
More recently, in the context of the Richer Event Description (RED) annotation scheme \citep{ikuta-etal-2014-challenges,ogorman2016red,ogorman2021red}, %
the annotation of events with finer-grained Vendler-style situation types has been proposed \citep{croft-etal-2016-annotation}.
\citet{falk-martin-2016-automatic} select 167 frequent French verbs from a lexical resource \citep{franccois2007presentation}, and label the corresponding 1199 entries (\enquote{readings}) with eight aspectual classes similar to those of \citet{mourelatos1978events} as listed in \aref{eventtypes}.
Their classification task is finer-grained than labeling verb types, but coarser-grained than clause-level labeling.
\citet{egg-etal-2019-annotation} annotate 4200 clauses from the German SdeWac \citep{faass2013sdewac} with the features stative, durative vs. punctual, and boundedness.
With the aim of improving zero-shot image to verb recognition, \citet{zellers-choi-2017-zero} crowd-source a dataset of 1710 verb templates (such as \enquote{put up}) for 1203 different verbs annotated with the four Vendler categories.
The Caption dataset \citep{alikhani-stone-2019-caption} annotates image captions with the features stative, durative, punctual, telic, and atelic.

\textbf{Modeling.}
\citet{zarcone-lenci-2008-computational} train a maximum entropy classifier that uses adverbial, morphological, and syntactic features, as well as features capturing argument structure.
\citet{richard2012lexical} uses an SVM \citep{cortes1995support} with features similar to the linguistic indicators of \citet{siegel-mckeown-2000-learning}.
\citet{zellers-choi-2017-zero} use GloVe embeddings \citep{pennington-etal-2014-glove} to represent the input verb.
They concatenate this embedding with a phrase embedding of the verb's dictionary definition computed using a recurrent neural network, and add a linear layer on top.
\citet{hermes2015automatic} induce Vendler classes for German verb types, using Siegel-style distributional features extracted from dependency-parsed corpus data (3000 sentences per verb type) and SVMs for type-level classification.
\citet{hermes2018supervised} compare the above framework to shallow distributional vectors considering only co-occurring word types for each verb.
Finally, \citet{gantt2022decomposing} propose a generative model for event types (and other information) fitted from 
English data annotated with a range of aspectual and other properties.
The resulting 4-class event type inventory closely resembles Vendler classes in the distinctions it draws.

The main obstacle to systematically comparing the works mentioned in this section is that their authors %
make differing choices for the granularity of both the annotation scheme and the computational modeling, and conclusions drawn from the studies depend on these choices.
For better comparability, we need better benchmarking.

\section{Grammatical Aspect / Viewpoint}
\label{sec:grammAspect}

Phenomena treated as \textit{grammatical aspect} or \textit{viewpoint} \citep{smith1997parameter} take different views (the entire situation vs. a part of it), or signal recurrence. %

\subsection{Perfective vs. Imperfective}
The \textit{perfective} viewpoint presents situations as complete with both an initial and a final endpoint, while the
\textit{imperfective} viewpoint makes only certain parts of the situation visible to the receiver \citep{smith1997parameter}.
The situation in \eeref{grammAspect}{a} is presented imperfectively, focusing on the middle phase of John's eating.
In contrast, \eeref{grammAspect}{b} is viewed in the perfective.
Here, the interpretation is that (b) happens during the interval at which (a) is true.

\begin{example}
(a) John was eating a sandwich \textit{(imperfective)}\\
\hspace*{7.1mm}(b) when Susan entered. \textit{(perfective)}
\label{ex:grammAspect}
\end{example}

The perfective/imperfective distinction is observed cross-linguistically, although some languages make even finer distinctions. For example, Chinese has two perfective aspect markers and two imperfective aspect markers. The two perfective markers \emph{le} (\ref{ex:zh-aspect}a) and \emph{guo} (\ref{ex:zh-aspect}b) differ in the present relevance, with the former indicating completion of a situation while the latter emphasizes the experience of having been through a situation. The two imperfective aspect markers \emph{\zheng4\zai4} and \emph{zhe} both indicate a situation is on-going at the reference time, but the latter emphasizes the resulting state of a situation \cite{chao1968grammar,li1989mandarin,ljungqvist2007guo}.

\begin{example}
\label{ex:zh-aspect}
(a) \ta1 \jin4 le \fang2\jian1 \enquote{\itshape He has entered the room\\ \hspace*{12mm} (and is still in the room)}\\ 
\hspace*{7mm}(b) \ta1 \jin4 \guo4 \fang2\jian1 \enquote{\itshape He entered the room \\ \hspace*{10mm} (at some point but is no longer in the room)}\\
\hspace*{7mm}(c) \ta1 \zheng4\zai4 \jin4 \fang2\jian1 \enquote{\itshape He is entering the \\ \hspace*{11mm} room}\\
\hspace*{7mm}(d) \fang2\jian1 de \men2 \kai2 zhe   \enquote{\itshape The door of the \\\hspace*{11mm} room is open.}
\end{example}

While also referring to endpoints, the concept of viewpoint differs from telicity: telicity describes types of situations independent of which phase is focused (e.g., \enquote{eating a sandwich} in \eeref{grammAspect}{a} is an (inherently-telic) accomplishment); viewpoint adds the focus (in this case on the phase during which the eating happens and the endpoint has not yet been reached; more details in Appendix~\ref{sec:viewpoint}).
It is the perfective-imperfective distinction that has traditionally been referred to as \textit{aspect} in Slavic linguistics (Filip 1999).
Computational work leverages parallel corpora to map the distinction, which is partially \textit{overt} (i.e., explicit in the morphosyntax) in Slavic languages, to English text \citep{stambolieva-2011-parallel,friedrich-gateva-2017-classification}.
Despite overall high accuracy, the models still struggle to reach high scores for the minority class \textit{atelic}.
In addition, such annotation projection approaches are not easily scalable, as they require strong knowledge of the languages involved.

\subsection{Habituality}
\label{sec:habituality}
Habituals such as \eeref{habitual}{b} are sentences that \enquote{express regularities about the world which constitute generalizations over events and activities} \citep{carlson2005generics}; on a sentence-level, they can be regarded as \enquote{derived statives} \citep{smith1997parameter}.
In contrast, the term \textit{episodic} refers to particular events. 
Habituals %
allow exceptions, e.g., \eeref{habitual}{b} is still true if Mary sometimes takes the train.

\begin{example}
(a) Mary cycled to work. \textit{(episodic)}\\
\hspace*{8mm} (b) Mary cycles to work. \textit{(habitual)}
\label{ex:habitual}
\end{example}

\noindent
Habitual sentences may also use stative predicates as in \eref{habitual-stative}, generalizing over situations in which some state applies \citep{smith2005aspectual}.
\begin{example}
Sloths sometimes rest on trees. \textit{(habitual)}
\label{ex:habitual-stative}
\end{example}

\noindent
Habituals are not restricted to what one would usually consider a matter of habit \citep{carlson2005generics}; they can also have inanimate subjects as in \eref{habitual-generic}.
\begin{example}
Glass breaks easily. \textit{(habitual)}
\label{ex:habitual-generic}
\end{example}

\noindent
Habituality is not to be confused with \textit{iterativity}, which states that a situation occurs repeatedly, but not regularly, as in \enquote{the light flashed.} %
Borderline cases are discussed in \aref{habituals-special-cases}.

\textbf{Data and Modeling.}
\citet{mathew2009supervised} randomly select 1052 sentences for 57 verbs from the Penn TreeBank \citep{marcus-etal-1993-building} and manually mark them as habitual or episodic.
They train both a decision tree and a Naive Bayes classifier on syntactic features extracted from gold parses.
\citet{friedrich-pinkal-2015-automatic} argue that in order to be able to apply such a model on free text, a third category, which they call \textit{static}, needs to be taken into consideration, which covers lexically stative, negated, and modalized clauses.
They experiment with Random Forest classifiers \citep{breiman2001random} using syntactic-semantic features and linguistic indicators (see \sref{stativity}).
\citet{govindarajan-etal-2019-decomposing} label UD-EWT \citep{silveira14gold} with real-valued properties indicating e.g., the \enquote{degree} of habituality.
Their multi-layer perceptron uses features from verb databases such as LCS and GLoVe embeddings.
Results for predicting habituality are overall promising, although there is a need for experiments with careful consideration of and controls over the verb types involved.

Other recent related work \citep{williams2012srw,williams-katz-2012-extracting,vempala-etal-2018-determining} extracts typical durations (in terms of actual time measures) for verb lemmas from Twitter.
They distinguish episodic and habitual uses of the verbs, using the method of \citet{mathew2009supervised}, and collect typical durations (e.g., \enquote{seconds} or \enquote{weeks}) for episodic and habitual uses separately for each verb.
The problem has further been studied in the context of acquiring common sense knowledge \citep[e.g.,][]{zhou-etal-2019-going,yang-etal-2020-improving}.

\subsection{Grammatical Aspect in Recent Syntactic-Semantic Frameworks}
\label{sec:grammAspect-frameworks}
The \textbf{Universal Dependencies} (UD) guidelines\footnote{\href{https://universaldependencies.org/u/feat/Aspect.html}{https://universaldependencies.org/u/feat/Aspect.html}} define six aspectual \textit{features} for verbs: Hab, Imp, Iter, Perf, Prog, and Prosp (prospective, for relative future).
The categories are assumed to be language-specific and close to the respective morphologies, and
the exact definition of each feature is left to the individual languages.
UD parsers frequently treat the identification of these features as a tagging task \citep[e.g.,][]{kondratyuk-straka-2019-75}.

In the context of a pilot study on integrating aspect into Abstract Meaning Representations \citep[][\textbf{AMR}]{banarescu-etal-2013-abstract},
\citet{donatelli-etal-2018-annotation} propose to indicate whether a clause is habitual or ongoing (in addition to marking stativity and telicity).
Tense and aspect annotation has been incorporated into Dialogue-AMR and used to annotate a corpus of human-robot interactions \cite{bonial-etal-2019-augmenting,bonial-etal-2020-dialogue}.
The Uniform Meaning Representations (\textbf{UMR}) framework \citep{vanGysel2021} uses a lattice for typologically-motivated annotation of aspect.
The lattice begins by making the distinction between three categories: \textsc{habitual}, which covers recurring states and events; \textsc{process}, which covers all non-recurring dynamic situations; and \textsc{imperfective}, designated for non-recurring states and atelic processes. 
The \textsc{process} category is further divided into ongoing \textsc{activities}, \textsc{endeavors} which have begun but not finished, and \textsc{performances}, which are completed dynamic processes.
See Appendix \ref{sec:umrAspect} for the full lattice. %

\citet{chen-etal-2021-autoaspect} build a rule-based system leveraging syntactic and semantic cues to annotate English sentences according to the UMR aspect lattice.
Due to the recency of these frameworks, so far there is  only very limited computational work.

\section{Situation Entities}
\label{sec:sitent}

In this section, we review work on automatically classifying \textit{situation entity} (SE) types (see inventory in Appendix \ref{sec:ses-full}).
SEs are \enquote{semantic concepts organized according to their internal temporal properties} \citep{smith2003modes}.
They are introduced to the discourse by a clause’s verb constellation, i.e., the clause’s main verb and its arguments and modifiers.
Deciding on the type of an SE thus involves the combination of lexical and syntactic factors.
In contrast to eventuality types (\sref{vendler}), situation entities capture the linguistic nature of the clause \citep{friedrich-palmer-2014-situation}.
\textsc{Event}s include all dynamic verb constructions referring to particular non-recurring situations, no matter whether there is a pre-defined endpoint or not. They may occur in the perfective or imperfective.
\textsc{Generalizing sentence}s correspond to habituals as described in \sref{habituality}, with the exception of \textsc{Generic Sentence}s that comprise all sentences making statements about kinds \citep{krifka1995genericity}.

\textbf{Datasets.}
\citet{palmer-etal-2007-sequencing} present 6065 clauses taken from the Brown corpus \citep{francis1979brown} and MUC-6 \citep{grishman1996message}, manually annotated with SE types.
The SitEnt dataset \citep{friedrich-etal-2016-situation,Friedrich_2017} consists of 40,000 clauses from 13 genres annotated for SE types with substantial agreement \citep{friedrich-palmer-2014-situation}.
\citet{govindarajan-etal-2019-decomposing} crowdsource UDW-G, a dataset of 37,146 arguments and 33,114 predicates in 16,222 sentences %
with continuous judgments on a scale from 1 to 5 with distinctions essentially following the schema of \citet{friedrich-etal-2016-situation}.

\textbf{Modeling.}
\citet{palmer-etal-2007-sequencing} use a maximum entropy sequence tagger with part-of-speech and CCG-based \citep{steedman2000syntactic} syntactic features to predict labels for each sentence separately. In an oracle experiment, they show that including the label of the previous sentence(s) as a feature improves performance.
\citet{friedrich-etal-2016-situation} present the first true sequence labeling approach to SE types classification using conditional random fields \citep{lafferty2001conditional} with distributional features in the form of Brown cluster IDs \citep{brown1992class} and syntactic-semantic features. %
\citet{kober-etal-2020-aspectuality} propose to classify predicational aspect with distributional semantics, using non-contextualized word embeddings and sums of the word vectors of the target words and their context words. %
Several works \citep{palmer-etal-2007-sequencing,palmer2014genre,friedrich-etal-2016-situation} find that sequence information mostly helps when training in-domain and has the biggest impact on identifying \textsc{Generic Sentences}, which often require discourse context for classification, even by human annotators \citep{friedrich-pinkal-2015-discourse}.

SE types have also been modeled using a variety of neural approaches.
\citet{becker-etal-2017-classifying} employ a GRU-based \citep{cho-etal-2014-properties} RNN with an attention mechanism.
\citet{dai-huang-2018-building} dynamically build context-aware clause representations, informed by their paragraph-wide contexts.
They propose a hierarchical recurrent neural network that reads entire paragraphs at once, learning interdependencies for clauses.
Their model uses word2vec \citep{mikolov2013efficient} embeddings for words, and first computes contextualized word embeddings using a BiLSTM over the entire paragraph.
Clause embeddings are formed by max pooling over the word embeddings of each clause, and then clause embeddings are further contextualized via another BiLSTM.
For further improving SE classification, they add a CRF layer on top, and finally predict SE labels via softmax.
Recently, BERT \citep{devlin-etal-2019-bert} and GPT-2 \citep{radford2019language} have been employed for the classification task by \citet{rezaee-etal-2021-discriminative}, who also compare to using ParBERT \citep{cohan-etal-2019-pretrained}.
While BERT only considers one clause at a time, ParBERT reads several sentences at once, separated by [SEP] tokens, and then uses the embeddings of the [SEP] tokens to predict a label for the corresponding clause.
Comparing a variety of neural models based on non-contextualized word embeddings on their situation-entity related regression tasks, \citet{govindarajan-etal-2019-decomposing} reach levels similar to human agreement.

The computational problem of identifying SE types has recently been studied extensively.
Overall, SE patterns are specific to the domain, genre or \textit{discourse mode} \citep{smith2003modes}.
Future research could use SE-style analysis for NLP tasks like temporal processing or information extraction.
\section{How Can We Move Forward?}
\label{sec:analysis}

In this section, we discuss potential steps that could lead to more successful modeling of aspectual information, as well as how to leverage that information in NLP applications.

\subsection{Dataset Construction}
To date, there is no consensus or standard across languages regarding where aspect lives.
Ongoing efforts in the UMR community aim to develop typologically-informed aspectual representations that work across languages, but so far very few UMR datasets exist.
Whether achieving a standard for aspectual representation is a necessary step or not, in order to make more rapid progress on the computational modeling of aspect, first the various models need to be \textbf{benchmarked} on the same tasks and datasets.
\textbf{Dataset construction} in this area is challenging.
Annotators not only need to have extensive linguistic training but also often domain expertise.
Formulating the problem as a crowdsourcing task is a possible direction \citep{govindarajan-etal-2019-decomposing}, but obtaining consistently high agreement is (as for most tasks) difficult in this setup.
\textbf{Multilingual} datasets and models are still under-researched,
yet they are of particular relevance due to applications in computer-aided language learning or the evaluation of machine translation (see examples in \aref{potential_applications}).

\subsection{Modeling}
More research is also needed on whether recent \textbf{embeddings} or text encoders capture or abstract away from aspectual features.
Recent experimental findings on the SitEnt dataset show that, when training data within the same domain and genre is available, performance of pre-trained transformers is in the range of human agreement \citep{metheniti-etal-2022-about,dai-huang-2018-building}.
The current state-of-the-art model based on BERT-GPT achieves a macro-average F1 of 79.1 compared to an estimate of human performance (78.6).
Yet, most verbs are either not ambiguous or have a strongly predominant majority class that the models memorize \citep{friedrich-palmer-2014-automatic,kober-etal-2020-aspectuality}.
We expect much insight to come from further studying verb types that behave differently depending on the context.
Moreover, results on cross-genre classification of SE types are yet inconclusive.
\citet{dai-huang-2018-building} report high cross-genre scores for BERT, ParBERT, and BERT+GPT, but out of these models, there is no clear winner across all genres.
Overall, when only using out-of-genre training data, performance of the BERT+GPT model still drops to an F1 of 70.7.

Investigating aspectual information in the context of \textbf{document-level models} \citep[e.g.,][]{longformer} is another opportunity: %
cues for aspectual interpretion may occur earlier in the discourse than the sentence being interpreted (see App. \ref{sec:potential_applications}).

Linguistic indicators \citep{siegel2000learning} have been shown to be useful for predicting aspectual features in the absence of training data.
Another research direction should hence look at how aspectual information can be \textbf{induced from raw data} or parallel corpora on a larger scale and for a broader set of distinctions.
Future work should also investigate the interaction with other elements of semantics such as argument structure.

In sum,
systems for predicting aspectual information encounter several challenges: (a) many verb types are ambiguous, (b) datasets for the task show significant class imbalance, and (c) for many verb types, no training data at all is available.
Initial experiments with careful controls for different groups of verbs were highly promising \citep{friedrich-palmer-2014-automatic,kober-etal-2020-aspectuality}, but available annotated data is sparse, especially for ambiguous verb types.
More research is required to disentangle the effects of potentially lexically biased datasets, class imbalance, and sparsity.

\subsection{Applications}
Automatic classifiers for aspectual distinctions clearly have \textbf{applications} in the \textbf{digital humanities} and quantitative linguistics, where research questions may directly target the use of aspect.
Recent highly fluent (but not necessarily always accurate) language generation models such as GPT-3 \citep{brown2020language} and ChatGPT\footnote{\href{https://openai.com/blog/chatgpt}{https://openai.com/blog/chatgpt}} open up a new level of writing assistance or language learning systems.
We believe that a fruitful direction for future research is to make such systems explainable, also in the sense that they are able to give linguistically founded qualitative feedback to the user about why an aspectual form is correct or wrong in a particular context.
Benchmarking on datasets explicitly annotated for aspectual information will let us estimate the degree to which text generated by the models is in line with the author's intentions.
In \aref{chat-gpt-logs}, we provide some chat logs of conversations with ChatGPT that illustrate how far it is from being an adequate tool for teaching the linguistic concepts described in this survey.

However, we argue that paying attention to aspect is necessary to arrive at fluent human-level NLP systems as motivated by our \enquote{miss the flight} example in \sref{intro}.
In \aref{potential_applications}, we discuss two potential application areas: temporal processing and machine translation.
In a nutshell, some works suggest that embedding spaces still struggle to capture the nature of tense and aspect, with consequences for commonsense reasoning driven by temporal properties, and can hence lead to wrong conclusions.
Similarly, machine translation systems may (and still do) go wrong when cues for the correct temporal form occur much earlier in the document then the sentence being translated.
More research is needed on how to \textbf{incorporate features} explicitly or \textbf{guide systems} implicitly towards the right output with regard to aspectual information, and also on how to convincingly evaluate generated output with regard to aspectual plausibility.

\section{Conclusion}
This paper provides an overview of the linguistic concepts and terminology associated with aspect, and also surveys existing datasets and prior work in computational modeling of aspect.
Because the expression of aspect varies across languages, the proper incorporation of aspect into downstream NLP tasks requires language-dependent research rooted in the language-independent categories surveyed here.
The semantic concepts expressed by lexical and grammatical aspect play essential roles in semantic interpretation, and their proper coding and evaluation must not be neglected if  we are to one day develop human-level NLU systems.
\newpage

\section*{Limitations}
This survey reviews linguistic and computational work on lexical and grammatical aspect.
While we believe that we covered the vast majority of relevant computational works, we of course only scratch the surface of the body of theoretical linguistic work and need to simplify at times.

As the term \textit{aspect} is highly ambiguous and used in NLP primarily within sentiment analysis, it is difficult to search for relevant papers.
Our search hence relied mainly on following citation networks.

\section*{Ethical Considerations}
We did not identify any potential ethical issues with this survey.

\section*{Acknowledgements}
We thank the anonymous reviewers for their extremely valuable comments.
This survey incorporates insights from countless extremely valuable discussions with many researchers and students.
In particular, we thank Manfred Pinkal, Josef van Genabith, Bonnie Webber, Mark Steedman, Mike Lewis, Omri Abend, Melissa Peate Sorensen, Michaela Regneri, Andrea Horbach, Stefan Thater, Vera Demberg, Ines Rehbein, Caroline Sporleder, Fatemeh Torabi Asr, Asad Sayeed, Alessandra Zarcone, Hanna Kermes, Damyana Gateva, Anna Nedoluzhko, Michal Novák, Martha Palmer, and Laura Michaelis.

\bibliography{anthology,custom}

\begin{thebibliography}{137}
\expandafter\ifx\csname natexlab\endcsname\relax\def\natexlab#1{#1}\fi

\bibitem[{Alikhani and Stone(2019)}]{alikhani-stone-2019-caption}
Malihe Alikhani and Matthew Stone. 2019.
\newblock \href {https://doi.org/10.18653/v1/W19-1806} {{``}caption{''} as a
  coherence relation: Evidence and implications}.
\newblock In \emph{Proceedings of the Second Workshop on Shortcomings in Vision
  and Language}, pages 58--67, Minneapolis, Minnesota. Association for
  Computational Linguistics.

\bibitem[{Banarescu et~al.(2013)Banarescu, Bonial, Cai, Georgescu, Griffitt,
  Hermjakob, Knight, Koehn, Palmer, and
  Schneider}]{banarescu-etal-2013-abstract}
Laura Banarescu, Claire Bonial, Shu Cai, Madalina Georgescu, Kira Griffitt, Ulf
  Hermjakob, Kevin Knight, Philipp Koehn, Martha Palmer, and Nathan Schneider.
  2013.
\newblock \href {https://aclanthology.org/W13-2322} {{A}bstract {M}eaning
  {R}epresentation for sembanking}.
\newblock In \emph{Proceedings of the 7th Linguistic Annotation Workshop and
  Interoperability with Discourse}, pages 178--186, Sofia, Bulgaria.
  Association for Computational Linguistics.

\bibitem[{Becker et~al.(2017)Becker, Staniek, Nastase, Palmer, and
  Frank}]{becker-etal-2017-classifying}
Maria Becker, Michael Staniek, Vivi Nastase, Alexis Palmer, and Anette Frank.
  2017.
\newblock \href {https://doi.org/10.18653/v1/S17-1027} {Classifying semantic
  clause types: Modeling context and genre characteristics with recurrent
  neural networks and attention}.
\newblock In \emph{Proceedings of the 6th Joint Conference on Lexical and
  Computational Semantics (*{SEM} 2017)}, pages 230--240, Vancouver, Canada.
  Association for Computational Linguistics.

\bibitem[{Beltagy et~al.(2020)Beltagy, Peters, and Cohan}]{longformer}
Iz~Beltagy, Matthew~E. Peters, and Arman Cohan. 2020.
\newblock \href {http://arxiv.org/abs/2004.05150} {Longformer: The
  long-document transformer}.
\newblock \emph{CoRR}, abs/2004.05150.

\bibitem[{Bender and Koller(2020)}]{bender-koller-2020-climbing}
Emily~M. Bender and Alexander Koller. 2020.
\newblock \href {https://doi.org/10.18653/v1/2020.acl-main.463} {Climbing
  towards {NLU}: {On} meaning, form, and understanding in the age of data}.
\newblock In \emph{Proceedings of the 58th Annual Meeting of the Association
  for Computational Linguistics}, pages 5185--5198, Online. Association for
  Computational Linguistics.

\bibitem[{Bethard(2013)}]{bethard-2013-cleartk}
Steven Bethard. 2013.
\newblock \href {https://aclanthology.org/S13-2002} {{C}lear{TK}-{T}ime{ML}: A
  minimalist approach to {T}emp{E}val 2013}.
\newblock In \emph{Second Joint Conference on Lexical and Computational
  Semantics (*{SEM}), Volume 2: Proceedings of the Seventh International
  Workshop on Semantic Evaluation ({S}em{E}val 2013)}, pages 10--14, Atlanta,
  Georgia, USA. Association for Computational Linguistics.

\bibitem[{Bethard and Martin(2006)}]{bethard-martin-2006-identification}
Steven Bethard and James~H. Martin. 2006.
\newblock \href {https://aclanthology.org/W06-1618} {Identification of event
  mentions and their semantic class}.
\newblock In \emph{Proceedings of the 2006 Conference on Empirical Methods in
  Natural Language Processing}, pages 146--154, Sydney, Australia. Association
  for Computational Linguistics.

\bibitem[{Bethard et~al.(2016)Bethard, Savova, Chen, Derczynski, Pustejovsky,
  and Verhagen}]{bethard-etal-2016-semeval}
Steven Bethard, Guergana Savova, Wei-Te Chen, Leon Derczynski, James
  Pustejovsky, and Marc Verhagen. 2016.
\newblock \href {https://doi.org/10.18653/v1/S16-1165} {{S}em{E}val-2016 task
  12: Clinical {T}emp{E}val}.
\newblock In \emph{Proceedings of the 10th International Workshop on Semantic
  Evaluation ({S}em{E}val-2016)}, pages 1052--1062, San Diego, California.
  Association for Computational Linguistics.

\bibitem[{Bethard et~al.(2017)Bethard, Savova, Palmer, and
  Pustejovsky}]{bethard-etal-2017-semeval}
Steven Bethard, Guergana Savova, Martha Palmer, and James Pustejovsky. 2017.
\newblock \href {https://doi.org/10.18653/v1/S17-2093} {{S}em{E}val-2017 task
  12: Clinical {T}emp{E}val}.
\newblock In \emph{Proceedings of the 11th International Workshop on Semantic
  Evaluation ({S}em{E}val-2017)}, pages 565--572, Vancouver, Canada.
  Association for Computational Linguistics.

\bibitem[{Bonial et~al.(2020)Bonial, Donatelli, Abrams, Lukin, Tratz, Marge,
  Artstein, Traum, and Voss}]{bonial-etal-2020-dialogue}
Claire Bonial, Lucia Donatelli, Mitchell Abrams, Stephanie~M. Lukin, Stephen
  Tratz, Matthew Marge, Ron Artstein, David Traum, and Clare Voss. 2020.
\newblock \href {https://aclanthology.org/2020.lrec-1.86} {Dialogue-{AMR}:
  {A}bstract {M}eaning {R}epresentation for dialogue}.
\newblock In \emph{Proceedings of the 12th Language Resources and Evaluation
  Conference}, pages 684--695, Marseille, France. European Language Resources
  Association.

\bibitem[{Bonial et~al.(2019)Bonial, Donatelli, Lukin, Tratz, Artstein, Traum,
  and Voss}]{bonial-etal-2019-augmenting}
Claire Bonial, Lucia Donatelli, Stephanie~M. Lukin, Stephen Tratz, Ron
  Artstein, David Traum, and Clare Voss. 2019.
\newblock \href {https://doi.org/10.18653/v1/W19-3322} {Augmenting {A}bstract
  {M}eaning {R}epresentation for human-robot dialogue}.
\newblock In \emph{Proceedings of the First International Workshop on Designing
  Meaning Representations}, pages 199--210, Florence, Italy. Association for
  Computational Linguistics.

\bibitem[{Bowman et~al.(2015)Bowman, Angeli, Potts, and
  Manning}]{bowman-etal-2015-large}
Samuel~R. Bowman, Gabor Angeli, Christopher Potts, and Christopher~D. Manning.
  2015.
\newblock \href {https://doi.org/10.18653/v1/D15-1075} {A large annotated
  corpus for learning natural language inference}.
\newblock In \emph{Proceedings of the 2015 Conference on Empirical Methods in
  Natural Language Processing}, pages 632--642, Lisbon, Portugal. Association
  for Computational Linguistics.

\bibitem[{Breiman(2001)}]{breiman2001random}
Leo Breiman. 2001.
\newblock Random forests.
\newblock \emph{Machine learning}, 45(1):5--32.

\bibitem[{Brennan et~al.(2013)Brennan, Schuhmann, and
  Batres}]{brennan2013entrainment}
Susan Brennan, Katharina Schuhmann, and Karla Batres. 2013.
\newblock Entrainment on the move and in the lab: The walking around corpus.
\newblock In \emph{Proceedings of the Annual Meeting of the Cognitive Science
  Society}, volume~35. Cognitive Science Society.

\bibitem[{Brent(1991{\natexlab{a}})}]{brent-1991-automatic}
Michael~R. Brent. 1991{\natexlab{a}}.
\newblock \href {https://aclanthology.org/E91-1039} {Automatic semantic
  classification of verbs from their syntactic contexts: An implemented
  classifier for stativity}.
\newblock In \emph{Fifth Conference of the {E}uropean Chapter of the
  Association for Computational Linguistics}, Berlin, Germany. Association for
  Computational Linguistics.

\bibitem[{Brent(1991{\natexlab{b}})}]{brent1991automatic}
Michael~R Brent. 1991{\natexlab{b}}.
\newblock Automatic semantic classification of verbs from their syntactic
  contexts: an implemented classifier for stativity.
\newblock In \emph{Proceedings of the fifth conference on European chapter of
  the Association for Computational Linguistics (ACL)}, pages 222--226, Berlin,
  Germany.

\bibitem[{Brown et~al.(1992)Brown, Desouza, Mercer, Pietra, and
  Lai}]{brown1992class}
Peter~F Brown, Peter~V Desouza, Robert~L Mercer, Vincent J~Della Pietra, and
  Jenifer~C Lai. 1992.
\newblock Class-based n-gram models of natural language.
\newblock \emph{Computational linguistics}, 18(4):467--479.

\bibitem[{Brown et~al.(2020)Brown, Mann, Ryder, Subbiah, Kaplan, Dhariwal,
  Neelakantan, Shyam, Sastry, Askell, Agarwal, Herbert-Voss, Krueger, Henighan,
  Child, Ramesh, Ziegler, Wu, Winter, Hesse, Chen, Sigler, Litwin, Gray, Chess,
  Clark, Berner, McCandlish, Radford, Sutskever, and
  Amodei}]{brown2020language}
Tom~B. Brown, Benjamin Mann, Nick Ryder, Melanie Subbiah, Jared Kaplan,
  Prafulla Dhariwal, Arvind Neelakantan, Pranav Shyam, Girish Sastry, Amanda
  Askell, Sandhini Agarwal, Ariel Herbert-Voss, Gretchen Krueger, Tom Henighan,
  Rewon Child, Aditya Ramesh, Daniel~M. Ziegler, Jeffrey Wu, Clemens Winter,
  Christopher Hesse, Mark Chen, Eric Sigler, Mateusz Litwin, Scott Gray,
  Benjamin Chess, Jack Clark, Christopher Berner, Sam McCandlish, Alec Radford,
  Ilya Sutskever, and Dario Amodei. 2020.
\newblock \href {http://arxiv.org/abs/2005.14165} {Language models are few-shot
  learners}.
\newblock \emph{cs.CL}.

\bibitem[{Carlson(2005)}]{carlson2005generics}
Gregory~N Carlson. 2005.
\newblock Generics, habituals and iteratives.
\newblock In Keith Brown, editor, \emph{The Encyclopedia of Language and
  Linguistics}. Elsevier Ltd.

\bibitem[{{\v{C}}erm{\'a}k and Rosen(2012)}]{ek2012case}
Franti {\v{C}}erm{\'a}k and Alexandr Rosen. 2012.
\newblock The case of intercorp, a multilingual parallel corpus.
\newblock \emph{International Journal of Corpus Linguistics}, 17(3):411--427.

\bibitem[{Chambers(2013)}]{chambers-2013-navytime}
Nathanael Chambers. 2013.
\newblock \href {https://aclanthology.org/S13-2012} {{N}avy{T}ime: Event and
  time ordering from raw text}.
\newblock In \emph{Second Joint Conference on Lexical and Computational
  Semantics (*{SEM}), Volume 2: Proceedings of the Seventh International
  Workshop on Semantic Evaluation ({S}em{E}val 2013)}, pages 73--77, Atlanta,
  Georgia, USA. Association for Computational Linguistics.

\bibitem[{Chao(1968)}]{chao1968grammar}
Yuen~Ren Chao. 1968.
\newblock \emph{A grammar of spoken Chinese}.
\newblock Univ of California Press.

\bibitem[{Chen and Palmer(2022)}]{chen-palmer-2022-contrast}
Daniel Chen and Alexis Palmer. 2022.
\newblock \href {https://aclanthology.org/2022.coling-1.354} {Contrast sets for
  stativity of {E}nglish verbs in context}.
\newblock In \emph{Proceedings of the 29th International Conference on
  Computational Linguistics}, pages 4028--4036, Gyeongju, Republic of Korea.
  International Committee on Computational Linguistics.

\bibitem[{Chen et~al.(2021)Chen, Palmer, and Vigus}]{chen-etal-2021-autoaspect}
Daniel Chen, Martha Palmer, and Meagan Vigus. 2021.
\newblock \href {https://doi.org/10.18653/v1/2021.law-1.4} {{A}uto{A}spect:
  Automatic annotation of tense and aspect for uniform meaning
  representations}.
\newblock In \emph{Proceedings of The Joint 15th Linguistic Annotation Workshop
  (LAW) and 3rd Designing Meaning Representations (DMR) Workshop}, pages
  36--45, Punta Cana, Dominican Republic. Association for Computational
  Linguistics.

\bibitem[{Cho et~al.(2014)Cho, van Merri{\"e}nboer, Bahdanau, and
  Bengio}]{cho-etal-2014-properties}
Kyunghyun Cho, Bart van Merri{\"e}nboer, Dzmitry Bahdanau, and Yoshua Bengio.
  2014.
\newblock \href {https://doi.org/10.3115/v1/W14-4012} {On the properties of
  neural machine translation: Encoder{--}decoder approaches}.
\newblock In \emph{Proceedings of {SSST}-8, Eighth Workshop on Syntax,
  Semantics and Structure in Statistical Translation}, pages 103--111, Doha,
  Qatar. Association for Computational Linguistics.

\bibitem[{Cohan et~al.(2019)Cohan, Beltagy, King, Dalvi, and
  Weld}]{cohan-etal-2019-pretrained}
Arman Cohan, Iz~Beltagy, Daniel King, Bhavana Dalvi, and Dan Weld. 2019.
\newblock \href {https://doi.org/10.18653/v1/D19-1383} {Pretrained language
  models for sequential sentence classification}.
\newblock In \emph{Proceedings of the 2019 Conference on Empirical Methods in
  Natural Language Processing and the 9th International Joint Conference on
  Natural Language Processing (EMNLP-IJCNLP)}, pages 3693--3699, Hong Kong,
  China. Association for Computational Linguistics.

\bibitem[{Comrie(1976)}]{comrie1976aspect}
Bernard Comrie. 1976.
\newblock \emph{Aspect: An introduction to the study of verbal aspect and
  related problems}, volume~2 of \emph{Cambridge Textbooks in Linguistics}.
\newblock Cambridge University Press.

\bibitem[{Cortes and Vapnik(1995)}]{cortes1995support}
Corinna Cortes and Vladimir Vapnik. 1995.
\newblock {Support-Vector Networks}.
\newblock \emph{Machine Learning}, 20(3):273--297.

\bibitem[{Costa and Branco(2012)}]{costa-branco-2012-aspectual}
Francisco Costa and Ant{\'o}nio Branco. 2012.
\newblock \href {https://aclanthology.org/E12-1027} {Aspectual type and
  temporal relation classification}.
\newblock In \emph{Proceedings of the 13th Conference of the {E}uropean Chapter
  of the Association for Computational Linguistics}, pages 266--275, Avignon,
  France. Association for Computational Linguistics.

\bibitem[{Croft et~al.(2016)Croft, Pe{\v{s}}kov{\'a}, and
  Regan}]{croft-etal-2016-annotation}
William Croft, Pavlina Pe{\v{s}}kov{\'a}, and Michael Regan. 2016.
\newblock \href {https://doi.org/10.18653/v1/W16-1002} {Annotation of causal
  and aspectual structure of events in {RED}: a preliminary report}.
\newblock In \emph{Proceedings of the Fourth Workshop on Events}, pages 8--17,
  San Diego, California. Association for Computational Linguistics.

\bibitem[{Dai and Huang(2018)}]{dai-huang-2018-building}
Zeyu Dai and Ruihong Huang. 2018.
\newblock \href {https://doi.org/10.18653/v1/D18-1368} {Building context-aware
  clause representations for situation entity type classification}.
\newblock In \emph{Proceedings of the 2018 Conference on Empirical Methods in
  Natural Language Processing}, pages 3305--3315, Brussels, Belgium.
  Association for Computational Linguistics.

\bibitem[{Devlin et~al.(2019)Devlin, Chang, Lee, and
  Toutanova}]{devlin-etal-2019-bert}
Jacob Devlin, Ming-Wei Chang, Kenton Lee, and Kristina Toutanova. 2019.
\newblock \href {https://doi.org/10.18653/v1/N19-1423} {{BERT}: Pre-training of
  deep bidirectional transformers for language understanding}.
\newblock In \emph{Proceedings of the 2019 Conference of the North {A}merican
  Chapter of the Association for Computational Linguistics: Human Language
  Technologies, Volume 1 (Long and Short Papers)}, pages 4171--4186,
  Minneapolis, Minnesota. Association for Computational Linguistics.

\bibitem[{Donatelli et~al.(2018)Donatelli, Regan, Croft, and
  Schneider}]{donatelli-etal-2018-annotation}
Lucia Donatelli, Michael Regan, William Croft, and Nathan Schneider. 2018.
\newblock \href {https://aclanthology.org/W18-4912} {Annotation of tense and
  aspect semantics for sentential {AMR}}.
\newblock In \emph{Proceedings of the Joint Workshop on Linguistic Annotation,
  Multiword Expressions and Constructions ({LAW}-{MWE}-{C}x{G}-2018)}, pages
  96--108, Santa Fe, New Mexico, USA. Association for Computational
  Linguistics.

\bibitem[{Dorr(2001)}]{dorr2001lcs}
Bonnie~J Dorr. 2001.
\newblock Lcs verb database, online software database of lexical conceptual
  structures and documentation, university of maryland.

\bibitem[{Dorr and Olsen(1997)}]{dorr-olsen-1997-deriving}
Bonnie~J. Dorr and Mari~Broman Olsen. 1997.
\newblock \href {https://doi.org/10.3115/976909.979637} {Deriving verbal and
  compositonal lexical aspect for {NLP} applications}.
\newblock In \emph{35th Annual Meeting of the Association for Computational
  Linguistics and 8th Conference of the {E}uropean Chapter of the Association
  for Computational Linguistics}, pages 151--158, Madrid, Spain. Association
  for Computational Linguistics.

\bibitem[{Dowty(1979)}]{dowty1979word}
David~R Dowty. 1979.
\newblock \emph{{Word meaning and Montague grammar -- The Semantics of Verbs
  and Times in Generative Semantics and in Montague's PTQ}}.
\newblock Studies in Linguistics and Philosophy. Springer.

\bibitem[{Dunietz et~al.(2020)Dunietz, Burnham, Bharadwaj, Rambow, Chu-Carroll,
  and Ferrucci}]{dunietz-etal-2020-test}
Jesse Dunietz, Greg Burnham, Akash Bharadwaj, Owen Rambow, Jennifer
  Chu-Carroll, and Dave Ferrucci. 2020.
\newblock \href {https://doi.org/10.18653/v1/2020.acl-main.701} {To test
  machine comprehension, start by defining comprehension}.
\newblock In \emph{Proceedings of the 58th Annual Meeting of the Association
  for Computational Linguistics}, pages 7839--7859, Online. Association for
  Computational Linguistics.

\bibitem[{Egg et~al.(2019)Egg, Prepens, and Roberts}]{egg-etal-2019-annotation}
Markus Egg, Helena Prepens, and Will Roberts. 2019.
\newblock \href {https://doi.org/10.18653/v1/P19-1323} {Annotation and
  automatic classification of aspectual categories}.
\newblock In \emph{Proceedings of the 57th Annual Meeting of the Association
  for Computational Linguistics}, pages 3335--3341, Florence, Italy.
  Association for Computational Linguistics.

\bibitem[{Faa{\ss} and Eckart(2013)}]{faass2013sdewac}
Gertrud Faa{\ss} and Kerstin Eckart. 2013.
\newblock Sdewac--a corpus of parsable sentences from the web.
\newblock In \emph{Language processing and knowledge in the Web}, pages 61--68.
  Springer.

\bibitem[{Falk and Martin(2016)}]{falk-martin-2016-automatic}
Ingrid Falk and Fabienne Martin. 2016.
\newblock \href {https://doi.org/10.18653/v1/S16-2002} {Automatic
  identification of aspectual classes across verbal readings}.
\newblock In \emph{Proceedings of the Fifth Joint Conference on Lexical and
  Computational Semantics}, pages 12--22, Berlin, Germany. Association for
  Computational Linguistics.

\bibitem[{Fellbaum(2010)}]{fellbaum2010wordnet}
Christiane Fellbaum. 2010.
\newblock Wordnet.
\newblock In \emph{Theory and applications of ontology: computer applications},
  pages 231--243. Springer.

\bibitem[{Filip(2012)}]{filip2012lexical}
Hana Filip. 2012.
\newblock Lexical aspect.
\newblock \emph{The Oxford handbook of tense and aspect}, pages 721--751.

\bibitem[{Filip and Carlson(1997)}]{filip1997sui}
Hana Filip and Gregory~N Carlson. 1997.
\newblock Sui generis genericity.
\newblock \emph{University of Pennsylvania Working Papers in Linguistics},
  4(2):7.

\bibitem[{Fillmore(1969)}]{fillmore1969types}
Charles~J Fillmore. 1969.
\newblock Types of lexical information.
\newblock In \emph{Studies in syntax and semantics}, pages 109--137. Springer.

\bibitem[{Francis and Ku\v{c}era(1979)}]{francis1979brown}
W~Nelson Francis and Henry Ku\v{c}era. 1979.
\newblock Brown corpus manual.
\newblock \emph{Brown University}.
\newblock Http://clu.uni.no/icame/brown/bcm.html.

\bibitem[{Fran{\c{c}}ois et~al.(2007)Fran{\c{c}}ois, Le~Pesant, and
  Leeman}]{franccois2007presentation}
Jacques Fran{\c{c}}ois, Denis Le~Pesant, and Danielle Leeman. 2007.
\newblock Pr{\'e}sentation de la classification des verbes fran{\c{c}}ais de
  jean dubois et fran{\c{c}}oise dubois-charlier.
\newblock \emph{Langue fran{\c{c}}aise}, 1:3--19.

\bibitem[{Friedrich and Gateva(2017)}]{friedrich-gateva-2017-classification}
Annemarie Friedrich and Damyana Gateva. 2017.
\newblock \href {https://doi.org/10.18653/v1/D17-1271} {Classification of
  telicity using cross-linguistic annotation projection}.
\newblock In \emph{Proceedings of the 2017 Conference on Empirical Methods in
  Natural Language Processing}, pages 2559--2565, Copenhagen, Denmark.
  Association for Computational Linguistics.

\bibitem[{Friedrich and
  Palmer(2014{\natexlab{a}})}]{friedrich-palmer-2014-automatic}
Annemarie Friedrich and Alexis Palmer. 2014{\natexlab{a}}.
\newblock \href {https://doi.org/10.3115/v1/P14-2085} {Automatic prediction of
  aspectual class of verbs in context}.
\newblock In \emph{Proceedings of the 52nd Annual Meeting of the Association
  for Computational Linguistics (Volume 2: Short Papers)}, pages 517--523,
  Baltimore, Maryland. Association for Computational Linguistics.

\bibitem[{Friedrich and
  Palmer(2014{\natexlab{b}})}]{friedrich-palmer-2014-situation}
Annemarie Friedrich and Alexis Palmer. 2014{\natexlab{b}}.
\newblock \href {https://doi.org/10.3115/v1/W14-4921} {Situation entity
  annotation}.
\newblock In \emph{Proceedings of {LAW} {VIII} - The 8th Linguistic Annotation
  Workshop}, pages 149--158, Dublin, Ireland. Association for Computational
  Linguistics and Dublin City University.

\bibitem[{Friedrich et~al.(2016)Friedrich, Palmer, and
  Pinkal}]{friedrich-etal-2016-situation}
Annemarie Friedrich, Alexis Palmer, and Manfred Pinkal. 2016.
\newblock \href {https://doi.org/10.18653/v1/P16-1166} {Situation entity types:
  automatic classification of clause-level aspect}.
\newblock In \emph{Proceedings of the 54th Annual Meeting of the Association
  for Computational Linguistics (Volume 1: Long Papers)}, pages 1757--1768,
  Berlin, Germany. Association for Computational Linguistics.

\bibitem[{Friedrich and
  Pinkal(2015{\natexlab{a}})}]{friedrich-pinkal-2015-automatic}
Annemarie Friedrich and Manfred Pinkal. 2015{\natexlab{a}}.
\newblock \href {https://doi.org/10.18653/v1/D15-1294} {Automatic recognition
  of habituals: a three-way classification of clausal aspect}.
\newblock In \emph{Proceedings of the 2015 Conference on Empirical Methods in
  Natural Language Processing}, pages 2471--2481, Lisbon, Portugal. Association
  for Computational Linguistics.

\bibitem[{Friedrich and
  Pinkal(2015{\natexlab{b}})}]{friedrich-pinkal-2015-discourse}
Annemarie Friedrich and Manfred Pinkal. 2015{\natexlab{b}}.
\newblock \href {https://doi.org/10.3115/v1/P15-1123} {Discourse-sensitive
  automatic identification of generic expressions}.
\newblock In \emph{Proceedings of the 53rd Annual Meeting of the Association
  for Computational Linguistics and the 7th International Joint Conference on
  Natural Language Processing (Volume 1: Long Papers)}, pages 1272--1281,
  Beijing, China. Association for Computational Linguistics.

\bibitem[{Friedrich(2017)}]{Friedrich_2017}
Annemarie~Silke Friedrich. 2017.
\newblock \href {https://doi.org/http://dx.doi.org/10.22028/D291-23666}
  {\emph{States, events, and generics: computational modeling of situation
  entity types}}.
\newblock Ph.D. thesis, Saarland University.

\bibitem[{Gantt et~al.(2022)Gantt, Glass, and White}]{gantt2022decomposing}
William Gantt, Lelia Glass, and Aaron~Steven White. 2022.
\newblock Decomposing and recomposing event structure.
\newblock \emph{Transactions of the Association for Computational Linguistics},
  10:17--34.

\bibitem[{Garey(1957)}]{garey1957verbal}
Howard~B Garey. 1957.
\newblock Verbal aspect in french.
\newblock \emph{Language}, 33(2):91--110.

\bibitem[{Govindarajan et~al.(2019)Govindarajan, Van~Durme, and
  White}]{govindarajan-etal-2019-decomposing}
Venkata Govindarajan, Benjamin Van~Durme, and Aaron~Steven White. 2019.
\newblock \href {https://doi.org/10.1162/tacl_a_00285} {Decomposing
  generalization: Models of generic, habitual, and episodic statements}.
\newblock \emph{Transactions of the Association for Computational Linguistics},
  7:501--517.

\bibitem[{Grishman and Sundheim(1996)}]{grishman1996message}
Ralph Grishman and Beth Sundheim. 1996.
\newblock Message understanding conference-6: A brief history.
\newblock In \emph{Proceedings of the 16th conference on Computational
  linguistics-Volume 1}, pages 466--471. Association for Computational
  Linguistics.

\bibitem[{Hacquard(2009)}]{hacquard2009interaction}
Valentine Hacquard. 2009.
\newblock On the interaction of aspect and modal auxiliaries.
\newblock \emph{Linguistics and Philosophy}, 32(3):279--315.

\bibitem[{Hermes et~al.(2015)Hermes, Richter, and
  Neuefeind}]{hermes2015automatic}
J\"{u}rgen Hermes, Michael Richter, and Claes Neuefeind. 2015.
\newblock \href
  {https://doi.org/https://konvens.org/proceedings/2015/GSCL-201518.pdf}
  {{Automatic Induction of German Aspectual Verb Classes in a Distributional
  Framework}}.
\newblock In \emph{Proceedings of the International Conference of the German
  Society for Computational Linguistics and Language Technology (GSCL)}, pages
  122--129, Duisburg-Essen, Germany.

\bibitem[{Hermes et~al.(2018)Hermes, Richter, and
  Neuefeind}]{hermes2018supervised}
J{\"u}rgen Hermes, Michael Richter, and Claes Neuefeind. 2018.
\newblock \href
  {https://doi.org/https://www.scitepress.org/papers/2018/67281/67281.pdf}
  {Supervised classification of aspectual verb classes in
  german-subcategorization-frame-based vs window-based approach: A comparison.}
\newblock In \emph{ICAART (2)}, pages 653--662.

\bibitem[{Ide et~al.(2008)Ide, Baker, Fellbaum, Fillmore, and
  Passonneau}]{ide-etal-2008-masc}
Nancy Ide, Collin Baker, Christiane Fellbaum, Charles Fillmore, and Rebecca
  Passonneau. 2008.
\newblock \href
  {http://www.lrec-conf.org/proceedings/lrec2008/pdf/617_paper.pdf} {{MASC}:
  the manually annotated sub-corpus of {A}merican {E}nglish}.
\newblock In \emph{Proceedings of the Sixth International Conference on
  Language Resources and Evaluation ({LREC}'08)}, Marrakech, Morocco. European
  Language Resources Association (ELRA).

\bibitem[{Ikuta et~al.(2014)Ikuta, Styler, Hamang, O{'}Gorman, and
  Palmer}]{ikuta-etal-2014-challenges}
Rei Ikuta, Will Styler, Mariah Hamang, Tim O{'}Gorman, and Martha Palmer. 2014.
\newblock \href {https://doi.org/10.3115/v1/W14-2903} {Challenges of adding
  causation to richer event descriptions}.
\newblock In \emph{Proceedings of the Second Workshop on {EVENTS}: Definition,
  Detection, Coreference, and Representation}, pages 12--20, Baltimore,
  Maryland, USA. Association for Computational Linguistics.

\bibitem[{Jung and Stent(2013)}]{jung-stent-2013-att1}
Hyuckchul Jung and Amanda Stent. 2013.
\newblock \href {https://aclanthology.org/S13-2004} {{ATT}1: Temporal
  annotation using big windows and rich syntactic and semantic features}.
\newblock In \emph{Second Joint Conference on Lexical and Computational
  Semantics (*{SEM}), Volume 2: Proceedings of the Seventh International
  Workshop on Semantic Evaluation ({S}em{E}val 2013)}, pages 20--24, Atlanta,
  Georgia, USA. Association for Computational Linguistics.

\bibitem[{Keelan(2012)}]{richard2012lexical}
Richard Keelan. 2012.
\newblock \emph{Lexical Aspectual Classification}.
\newblock University of Ottawa (Canada).

\bibitem[{Kenny(1963)}]{kenny1963action}
Anthony Kenny. 1963.
\newblock \emph{Action, emotion and will}.
\newblock Routledge.

\bibitem[{Klavans and
  Chodorow(1992{\natexlab{a}})}]{klavans-chodorow-1992-degrees}
Judith~L. Klavans and Martin Chodorow. 1992{\natexlab{a}}.
\newblock \href {https://aclanthology.org/C92-4177} {Degrees of stativity: The
  lexical representation of verb aspect}.
\newblock In \emph{{COLING} 1992 Volume 4: The 14th {I}nternational
  {C}onference on {C}omputational {L}inguistics}.

\bibitem[{Klavans and Chodorow(1992{\natexlab{b}})}]{klavans1992degrees}
Judith~L Klavans and Martin Chodorow. 1992{\natexlab{b}}.
\newblock Degrees of stativity: the lexical representation of verb aspect.
\newblock In \emph{Proceedings of the 14th conference on Computational
  linguistics (Coling)}, volume~4, pages 1126--1131, Nantes, France.

\bibitem[{Kober et~al.(2020)Kober, Alikhani, Stone, and
  Steedman}]{kober-etal-2020-aspectuality}
Thomas Kober, Malihe Alikhani, Matthew Stone, and Mark Steedman. 2020.
\newblock \href {https://doi.org/10.18653/v1/2020.coling-main.401}
  {Aspectuality across genre: A distributional semantics approach}.
\newblock In \emph{Proceedings of the 28th International Conference on
  Computational Linguistics}, pages 4546--4562, Barcelona, Spain (Online).
  International Committee on Computational Linguistics.

\bibitem[{Kober et~al.(2021)Kober, Weeds, Bertolini, and
  Weir}]{kober-etal-2021-data}
Thomas Kober, Julie Weeds, Lorenzo Bertolini, and David Weir. 2021.
\newblock \href {https://doi.org/10.18653/v1/2021.eacl-main.89} {Data
  augmentation for hypernymy detection}.
\newblock In \emph{Proceedings of the 16th Conference of the European Chapter
  of the Association for Computational Linguistics: Main Volume}, pages
  1034--1048, Online. Association for Computational Linguistics.

\bibitem[{Kondratyuk and Straka(2019)}]{kondratyuk-straka-2019-75}
Dan Kondratyuk and Milan Straka. 2019.
\newblock \href {https://doi.org/10.18653/v1/D19-1279} {75 languages, 1 model:
  Parsing {U}niversal {D}ependencies universally}.
\newblock In \emph{Proceedings of the 2019 Conference on Empirical Methods in
  Natural Language Processing and the 9th International Joint Conference on
  Natural Language Processing (EMNLP-IJCNLP)}, pages 2779--2795, Hong Kong,
  China. Association for Computational Linguistics.

\bibitem[{Krifka et~al.(1995)Krifka, Pelletier, Carlson, ter Meulen, Link, and
  Chierchia}]{krifka1995genericity}
Manfred Krifka, Francis~Jeffrey Pelletier, Gregory~N. Carlson, Alice ter
  Meulen, Godehard Link, and Gennaro Chierchia. 1995.
\newblock {Genericity: An Introduction}.
\newblock In Gregory~N. Carlson and Francis~Jeffry Pelletier, editors,
  \emph{The Generic Book}, Studies in Communication, Media, and Public Opinion,
  pages 1--124. University Of Chicago Press.

\bibitem[{Lafferty et~al.(2001)Lafferty, McCallum, and
  Pereira}]{lafferty2001conditional}
John Lafferty, Andrew McCallum, and Fernando Pereira. 2001.
\newblock \href {https://doi.org/https://repository.upenn.edu/cis_papers/159/}
  {Conditional random fields: Probabilistic models for segmenting and labeling
  sequence data}.
\newblock In \emph{Proceedings of the eighteenth International Conference on
  Machine Learning (ICML)}, volume~1, pages 282--289, Williamstown,
  Massachussetts.

\bibitem[{Lan et~al.(2019)Lan, Chen, Goodman, Gimpel, Sharma, and
  Soricut}]{albert}
Zhenzhong Lan, Mingda Chen, Sebastian Goodman, Kevin Gimpel, Piyush Sharma, and
  Radu Soricut. 2019.
\newblock \href {http://arxiv.org/abs/1909.11942} {{ALBERT:} {A} lite {BERT}
  for self-supervised learning of language representations}.
\newblock \emph{CoRR}, abs/1909.11942.

\bibitem[{Lascarides(1991)}]{lascarides1991progressive}
Alex Lascarides. 1991.
\newblock {The Progressive and the Imperfective Paradox}.
\newblock \emph{Synthese}, 87(3):401--447.

\bibitem[{Leech(1971)}]{leech1971meaning}
Geoffrey~N Leech. 1971.
\newblock \emph{Meaning and the English verb}.
\newblock Pearson Education.

\bibitem[{Li and Thompson(1989)}]{li1989mandarin}
Charles~N Li and Sandra~A Thompson. 1989.
\newblock \emph{Mandarin Chinese: A functional reference grammar}, volume~3.
\newblock Univ of California Press.

\bibitem[{Liu et~al.(2019)Liu, Ott, Goyal, Du, Joshi, Chen, Levy, Lewis,
  Zettlemoyer, and Stoyanov}]{roberta}
Yinhan Liu, Myle Ott, Naman Goyal, Jingfei Du, Mandar Joshi, Danqi Chen, Omer
  Levy, Mike Lewis, Luke Zettlemoyer, and Veselin Stoyanov. 2019.
\newblock \href {http://arxiv.org/abs/1907.11692} {Roberta: {A} robustly
  optimized {BERT} pretraining approach}.
\newblock \emph{CoRR}, abs/1907.11692.

\bibitem[{Ljungqvist(2007)}]{ljungqvist2007guo}
Marita Ljungqvist. 2007.
\newblock Le, guo and zhe in mandarin chinese: a relevance-theoretic account.
\newblock \emph{Journal of East Asian Linguistics}, 16(3):193--235.

\bibitem[{Llorens et~al.(2010)Llorens, Saquete, and
  Navarro-Colorado}]{llorens-etal-2010-timeml}
Hector Llorens, Estela Saquete, and Borja Navarro-Colorado. 2010.
\newblock \href {https://aclanthology.org/C10-1082} {{T}ime{ML} events
  recognition and classification: Learning {CRF} models with semantic roles}.
\newblock In \emph{Proceedings of the 23rd International Conference on
  Computational Linguistics (Coling 2010)}, pages 725--733, Beijing, China.
  Coling 2010 Organizing Committee.

\bibitem[{Lo{\'a}iciga and Grisot(2016)}]{loaiciga-grisot-2016-predicting}
Sharid Lo{\'a}iciga and Cristina Grisot. 2016.
\newblock \href {https://aclanthology.org/2016.lilt-13.3} {Predicting and using
  a pragmatic component of lexical aspect of simple past verbal tenses for
  improving {E}nglish-to-{F}rench machine translation}.
\newblock In \emph{Linguistic Issues in Language Technology, Volume 13, 2016}.
  CSLI Publications.

\bibitem[{Marcus et~al.(1993)Marcus, Santorini, and
  Marcinkiewicz}]{marcus-etal-1993-building}
Mitchell~P. Marcus, Beatrice Santorini, and Mary~Ann Marcinkiewicz. 1993.
\newblock \href {https://aclanthology.org/J93-2004} {Building a large annotated
  corpus of {E}nglish: The {P}enn {T}reebank}.
\newblock \emph{Computational Linguistics}, 19(2):313--330.

\bibitem[{Mathew and Katz(2009)}]{mathew2009supervised}
Thomas~A. Mathew and E.~Graham Katz. 2009.
\newblock {Supervised Categorization of Habitual and Episodic Sentences}.
\newblock In \emph{Sixth Midwest Computational Linguistics Colloquium},
  Bloomington, Indiana: Indiana University.

\bibitem[{Metheniti(2022)}]{metheniti-etal-2022-about}
Eleni Metheniti. 2022.
\newblock \href {https://doi.org/https://openreview.net/pdf?id=Bduz_QYg8Z5}
  {About time: Do transformers learn temporal verbal aspect?}
\newblock In \emph{Cognitive Modeling and Computational Linguistics (CMCL)},
  Dublin,Ireland.

\bibitem[{Metheniti et~al.(2021)Metheniti, van~de Cruys, and
  Hathout}]{metheniti-etal-2021-predire}
Eleni Metheniti, Tim van~de Cruys, and Nabil Hathout. 2021.
\newblock \href {https://aclanthology.org/2021.jeptalnrecital-taln.20}
  {Pr{\'e}dire l{'}aspect linguistique en anglais au moyen de transformers
  (classifying linguistic aspect in {E}nglish with transformers )}.
\newblock In \emph{Actes de la 28e Conf{\'e}rence sur le Traitement Automatique
  des Langues Naturelles. Volume 1 : conf{\'e}rence principale}, pages
  209--218, Lille, France. ATALA.

\bibitem[{Mikolov et~al.(2013)Mikolov, Chen, Corrado, and
  Dean}]{mikolov2013efficient}
Tomas Mikolov, Kai Chen, Greg Corrado, and Jeffrey Dean. 2013.
\newblock Efficient estimation of word representations in vector space.
\newblock \emph{arXiv preprint arXiv:1301.3781}.

\bibitem[{Moens and Steedman(1988)}]{moens-steedman-1988-temporal}
Marc Moens and Mark Steedman. 1988.
\newblock \href {https://aclanthology.org/J88-2003} {Temporal ontology and
  temporal reference}.
\newblock \emph{Computational Linguistics}, 14(2):15--28.

\bibitem[{Mourelatos(1978)}]{mourelatos1978events}
Alexander~PD Mourelatos. 1978.
\newblock Events, processes, and states.
\newblock \emph{Linguistics and philosophy}, 2(3):415--434.

\bibitem[{Myers and Palmer(2019)}]{myers-palmer-2019-cleartac}
Skatje Myers and Martha Palmer. 2019.
\newblock \href {https://doi.org/10.18653/v1/W19-3315} {{C}lear{TAC}: Verb
  tense, aspect, and form classification using neural nets}.
\newblock In \emph{Proceedings of the First International Workshop on Designing
  Meaning Representations}, pages 136--140, Florence, Italy. Association for
  Computational Linguistics.

\bibitem[{O'Gorman et~al.(2021)O'Gorman, Wright-Bettner, and
  Palmer}]{ogorman2021red}
Tim O'Gorman, Kristin Wright-Bettner, and Martha Palmer. 2021.
\newblock {The Richer Event Description Corpus for Event-Event Relations}.
\newblock \emph{Computational Analysis of Storylines: Making Sense of Events
  (Studies in Natural Language Processing)}.

\bibitem[{O’Gorman et~al.(2016)O’Gorman, Wright-Bettner, and
  Palmer}]{ogorman2016red}
Tim O’Gorman, Kristin Wright-Bettner, and Martha Palmer. 2016.
\newblock Richer event description: Integrating event coreference with
  temporal, causal and bridging annotation.
\newblock In \emph{Proceedings of the 2nd Workshop on Computing News Storylines
  (CNS 2016)}, pages 47--56.

\bibitem[{Palmer and Friedrich(2014)}]{palmer2014genre}
Alexis Palmer and Annemarie Friedrich. 2014.
\newblock {Genre distinctions and discourse modes: Text types differ in their
  situation type distributions}.
\newblock In \emph{Proceedings of the Symposium on Frontiers and Connections
  between Argumentation Mining and Natural Language Processing}, Bertinoro,
  Italy.

\bibitem[{Palmer et~al.(2007)Palmer, Ponvert, Baldridge, and
  Smith}]{palmer-etal-2007-sequencing}
Alexis Palmer, Elias Ponvert, Jason Baldridge, and Carlota Smith. 2007.
\newblock \href {https://aclanthology.org/P07-1113} {A sequencing model for
  situation entity classification}.
\newblock In \emph{Proceedings of the 45th Annual Meeting of the Association of
  Computational Linguistics}, pages 896--903, Prague, Czech Republic.
  Association for Computational Linguistics.

\bibitem[{Passonneau(1988{\natexlab{a}})}]{passonneau-1988-computational}
Rebecca~J. Passonneau. 1988{\natexlab{a}}.
\newblock \href {https://aclanthology.org/J88-2005} {A computational model of
  the semantics of tense and aspect}.
\newblock \emph{Computational Linguistics}, 14(2):44--60.

\bibitem[{Passonneau(1988{\natexlab{b}})}]{passonneau1988computational}
Rebecca~J Passonneau. 1988{\natexlab{b}}.
\newblock A computational model of the semantics of tense and aspect.
\newblock \emph{Computational Linguistics}, 14(2):44--60.

\bibitem[{Pennington et~al.(2014)Pennington, Socher, and
  Manning}]{pennington-etal-2014-glove}
Jeffrey Pennington, Richard Socher, and Christopher Manning. 2014.
\newblock \href {https://doi.org/10.3115/v1/D14-1162} {{G}lo{V}e: Global
  vectors for word representation}.
\newblock In \emph{Proceedings of the 2014 Conference on Empirical Methods in
  Natural Language Processing ({EMNLP})}, pages 1532--1543, Doha, Qatar.
  Association for Computational Linguistics.

\bibitem[{Poliak et~al.(2018)Poliak, Haldar, Rudinger, Hu, Pavlick, White, and
  Van~Durme}]{poliak-etal-2018-collecting}
Adam Poliak, Aparajita Haldar, Rachel Rudinger, J.~Edward Hu, Ellie Pavlick,
  Aaron~Steven White, and Benjamin Van~Durme. 2018.
\newblock \href {https://doi.org/10.18653/v1/D18-1007} {Collecting diverse
  natural language inference problems for sentence representation evaluation}.
\newblock In \emph{Proceedings of the 2018 Conference on Empirical Methods in
  Natural Language Processing}, pages 67--81, Brussels, Belgium. Association
  for Computational Linguistics.

\bibitem[{Pustejovsky et~al.(2003)Pustejovsky, Castano, Ingria, Sauri,
  Gaizauskas, Setzer, Katz, and Radev}]{pustejovsky2003timeml}
James Pustejovsky, Jos{\'e}~M Castano, Robert Ingria, Roser Sauri, Robert~J
  Gaizauskas, Andrea Setzer, Graham Katz, and Dragomir~R Radev. 2003.
\newblock Timeml: Robust specification of event and temporal expressions in
  text.
\newblock \emph{New directions in question answering}, 3:28--34.

\bibitem[{Pustejovsky et~al.(2010)Pustejovsky, Lee, Bunt, and
  Romary}]{pustejovsky-etal-2010-iso}
James Pustejovsky, Kiyong Lee, Harry Bunt, and Laurent Romary. 2010.
\newblock \href
  {http://www.lrec-conf.org/proceedings/lrec2010/pdf/55_Paper.pdf}
  {{ISO}-{T}ime{ML}: An international standard for semantic annotation}.
\newblock In \emph{Proceedings of the Seventh International Conference on
  Language Resources and Evaluation ({LREC}'10)}, Valletta, Malta. European
  Language Resources Association (ELRA).

\bibitem[{Radford et~al.(2019)Radford, Wu, Child, Luan, Amodei, Sutskever
  et~al.}]{radford2019language}
Alec Radford, Jeffrey Wu, Rewon Child, David Luan, Dario Amodei, Ilya
  Sutskever, et~al. 2019.
\newblock Language models are unsupervised multitask learners.
\newblock \emph{OpenAI blog}, 1(8):9.

\bibitem[{Ramm et~al.(2017)Ramm, Lo{\'a}iciga, Friedrich, and
  Fraser}]{ramm-etal-2017-annotating}
Anita Ramm, Sharid Lo{\'a}iciga, Annemarie Friedrich, and Alexander Fraser.
  2017.
\newblock \href {https://aclanthology.org/P17-4001} {Annotating tense, mood and
  voice for {E}nglish, {F}rench and {G}erman}.
\newblock In \emph{Proceedings of {ACL} 2017, System Demonstrations}, pages
  1--6, Vancouver, Canada. Association for Computational Linguistics.

\bibitem[{Reichart and Rappoport(2010)}]{reichart-rappoport-2010-tense}
Roi Reichart and Ari Rappoport. 2010.
\newblock \href {https://aclanthology.org/D10-1032} {Tense sense
  disambiguation: A new syntactic polysemy task}.
\newblock In \emph{Proceedings of the 2010 Conference on Empirical Methods in
  Natural Language Processing}, pages 325--334, Cambridge, MA. Association for
  Computational Linguistics.

\bibitem[{Rezaee et~al.(2021)Rezaee, Darvish, Kebe, and
  Ferraro}]{rezaee-etal-2021-discriminative}
Mehdi Rezaee, Kasra Darvish, Gaoussou~Youssouf Kebe, and Francis Ferraro. 2021.
\newblock \href {http://arxiv.org/abs/2109.07434} {Discriminative and
  generative transformer-based models for situation entity classification}.
\newblock \emph{CoRR}, abs/2109.07434.

\bibitem[{Rosen and Vav{\v{r}}{\'\i}n(2012)}]{rosen-vavrin-2012-building}
Alexandr Rosen and Martin Vav{\v{r}}{\'\i}n. 2012.
\newblock \href
  {http://www.lrec-conf.org/proceedings/lrec2012/pdf/200_Paper.pdf} {Building a
  multilingual parallel corpus for human users}.
\newblock In \emph{Proceedings of the Eighth International Conference on
  Language Resources and Evaluation ({LREC}'12)}, pages 2447--2452, Istanbul,
  Turkey. European Language Resources Association (ELRA).

\bibitem[{Saur{\'\i} et~al.(2005)Saur{\'\i}, Knippen, Verhagen, and
  Pustejovsky}]{sauri-etal-2005-evita}
Roser Saur{\'\i}, Robert Knippen, Marc Verhagen, and James Pustejovsky. 2005.
\newblock \href {https://aclanthology.org/H05-1088} {{E}vita: A robust event
  recognizer for {QA} systems}.
\newblock In \emph{Proceedings of Human Language Technology Conference and
  Conference on Empirical Methods in Natural Language Processing}, pages
  700--707, Vancouver, British Columbia, Canada. Association for Computational
  Linguistics.

\bibitem[{Siegel(1997)}]{siegel-1997-learning}
Eric~V. Siegel. 1997.
\newblock \href {https://aclanthology.org/W97-0318} {Learning methods for
  combining linguistic indicators to classify verbs}.
\newblock In \emph{Second Conference on Empirical Methods in Natural Language
  Processing}.

\bibitem[{Siegel(1999)}]{siegel-1999-corpus}
Eric~V. Siegel. 1999.
\newblock \href {https://doi.org/10.3115/1034678.1034704} {Corpus-based
  linguistic indicators for aspectual classification}.
\newblock In \emph{Proceedings of the 37th Annual Meeting of the Association
  for Computational Linguistics}, pages 112--119, College Park, Maryland, USA.
  Association for Computational Linguistics.

\bibitem[{Siegel and McKeown(1996)}]{siegel1996gathering}
Eric~V Siegel and Kathleen~R McKeown. 1996.
\newblock \href {https://doi.org/https://arxiv.org/pdf/cmp-lg/9610002.pdf}
  {Gathering statistics to aspectually classify sentences with a genetic
  algorithm}.
\newblock In \emph{Proceedings of the Second International Conference on New
  Methods in Language Processing}, Ankara, Turkey.

\bibitem[{Siegel and McKeown(2000{\natexlab{a}})}]{siegel2000learning}
Eric~V Siegel and Kathleen~R McKeown. 2000{\natexlab{a}}.
\newblock Learning methods to combine linguistic indicators: Improving
  aspectual classification and revealing linguistic insights.
\newblock \emph{Computational Linguistics}, 26(4):595--628.

\bibitem[{Siegel and
  McKeown(2000{\natexlab{b}})}]{siegel-mckeown-2000-learning}
Eric~V. Siegel and Kathleen~R. McKeown. 2000{\natexlab{b}}.
\newblock \href {https://aclanthology.org/J00-4004} {Learning methods to
  combine linguistic indicators:improving aspectual classification and
  revealing linguistic insights}.
\newblock \emph{Computational Linguistics}, 26(4):595--627.

\bibitem[{Silveira et~al.(2014)Silveira, Dozat, de~Marneffe, Bowman, Connor,
  Bauer, and Manning}]{silveira14gold}
Natalia Silveira, Timothy Dozat, Marie-Catherine de~Marneffe, Samuel Bowman,
  Miriam Connor, John Bauer, and Christopher~D. Manning. 2014.
\newblock A gold standard dependency corpus for {E}nglish.
\newblock In \emph{Proceedings of the Ninth International Conference on
  Language Resources and Evaluation (LREC-2014)}.

\bibitem[{Smith(1997)}]{smith1997parameter}
Carlota~S Smith. 1997.
\newblock \emph{The parameter of aspect}, volume~43 of \emph{Studies in
  Linguistics and Philosophy}.
\newblock Springer Science \& Business Media.

\bibitem[{Smith(2003)}]{smith2003modes}
Carlota~S Smith. 2003.
\newblock \emph{Modes of discourse: The local structure of texts}, volume 103.
\newblock Cambridge University Press.

\bibitem[{Smith(2005)}]{smith2005aspectual}
Carlota~S Smith. 2005.
\newblock Aspectual entities and tense in discourse.
\newblock In \emph{Aspectual inquiries}, pages 223--237. Springer.

\bibitem[{Stambolieva(2011)}]{stambolieva-2011-parallel}
Maria Stambolieva. 2011.
\newblock \href {https://aclanthology.org/W11-4306} {Parallel corpora in
  aspectual studies of non-aspect languages}.
\newblock In \emph{Proceedings of The Second Workshop on Annotation and
  Exploitation of Parallel Corpora}, pages 39--42, Hissar, Bulgaria.
  Association for Computational Linguistics.

\bibitem[{Steedman(2000)}]{steedman2000syntactic}
Mark Steedman. 2000.
\newblock \emph{The syntactic process}, volume~24 of \emph{Language, Speech,
  and Communication}.
\newblock MIT Press.

\bibitem[{Sun et~al.(2021)Sun, Liu, Qiu, and Huang}]{sun2021paradigmshift}
Tianxiang Sun, Xiangyang Liu, Xipeng Qiu, and Xuanjing Huang. 2021.
\newblock Paradigm shift in natural language processing.
\newblock \emph{arXiv preprint arXiv:2109.12575}.

\bibitem[{Thater et~al.(2010)Thater, F{\"u}rstenau, and
  Pinkal}]{thater-etal-2010-contextualizing}
Stefan Thater, Hagen F{\"u}rstenau, and Manfred Pinkal. 2010.
\newblock \href {https://aclanthology.org/P10-1097} {Contextualizing semantic
  representations using syntactically enriched vector models}.
\newblock In \emph{Proceedings of the 48th Annual Meeting of the Association
  for Computational Linguistics}, pages 948--957, Uppsala, Sweden. Association
  for Computational Linguistics.

\bibitem[{Trott et~al.(2020)Trott, Torrent, Chang, and
  Schneider}]{trott-etal-2020-construing}
Sean Trott, Tiago~Timponi Torrent, Nancy Chang, and Nathan Schneider. 2020.
\newblock \href {https://doi.org/10.18653/v1/2020.acl-main.462}
  {({R}e)construing meaning in {NLP}}.
\newblock In \emph{Proceedings of the 58th Annual Meeting of the Association
  for Computational Linguistics}, pages 5170--5184, Online. Association for
  Computational Linguistics.

\bibitem[{UzZaman et~al.(2013)UzZaman, Llorens, Derczynski, Allen, Verhagen,
  and Pustejovsky}]{uzzaman2013semeval}
Naushadand UzZaman, Hectorand Llorens, Leon Derczynski, James Allen, Marc
  Verhagen, and James Pustejovsky. 2013.
\newblock Semeval-2013 task 1: Tempeval-3: Evaluating time expressions, events,
  and temporal relations.
\newblock In \emph{Second Joint Conference on Lexical and Computational
  Semantics (*SEM), Volume 2: Proceedings of the Seventh International Workshop
  on Semantic Evaluation (SemEval 2013)}, pages 1--9, Atlanta, Georgia.

\bibitem[{Van~Gysel et~al.(2022)Van~Gysel, Vigus, Zhao, and Xue}]{vangysel2022}
Jens Van~Gysel, Meagan Vigus, Jin Zhao, and Nianwen Xue. 2022.
\newblock Developing a uniform meaning representation for natural language
  processing.
\newblock In \emph{LREC 2022 tutorial}.

\bibitem[{Van~Gysel et~al.(2021)Van~Gysel, Vigus, Chun, Lai, Moeller, Yao,
  O’Gorman, Cowell, Croft, Huang, Hajič, Martin, Oepen, Palmer, Pustejovsky,
  Vallejos, and Xue}]{vanGysel2021}
Jens E.~L. Van~Gysel, Meagan Vigus, Jayeol Chun, Kenneth Lai, Sarah Moeller,
  Jiarui Yao, Tim O’Gorman, Andrew Cowell, William Croft, Chu-Ren Huang, Jan
  Hajič, James~H. Martin, Stephan Oepen, Martha Palmer, James Pustejovsky,
  Rosa Vallejos, and Nianwen Xue. 2021.
\newblock \href {https://doi.org/10.1007/s13218-021-00722-w} {Designing a
  uniform meaning representation for natural language processing}.
\newblock \emph{KI - Künstliche Intelligenz}, 35(0):343--360.

\bibitem[{Vanmassenhove et~al.(2017)Vanmassenhove, Du, and
  Way}]{vanmassenhove2017investigating}
Eva Vanmassenhove, Jinhua Du, and Andy Way. 2017.
\newblock \href
  {https://doi.org/https://www.clips.uantwerpen.be/clinjdraft/clinj/article/view/73}
  {Investigating ‘aspect’in nmt and smt: Translating the english simple
  past and present perfect}.
\newblock \emph{Computational Linguistics in the Netherlands Journal},
  7:109--128.

\bibitem[{Vashishtha et~al.(2020)Vashishtha, Poliak, Lal, Van~Durme, and
  White}]{vashishtha-etal-2020-temporal}
Siddharth Vashishtha, Adam Poliak, Yash~Kumar Lal, Benjamin Van~Durme, and
  Aaron~Steven White. 2020.
\newblock \href {https://doi.org/10.18653/v1/2020.findings-emnlp.363} {Temporal
  reasoning in natural language inference}.
\newblock In \emph{Findings of the Association for Computational Linguistics:
  EMNLP 2020}, pages 4070--4078, Online. Association for Computational
  Linguistics.

\bibitem[{Vempala et~al.(2018)Vempala, Blanco, and
  Palmer}]{vempala-etal-2018-determining}
Alakananda Vempala, Eduardo Blanco, and Alexis Palmer. 2018.
\newblock \href {https://doi.org/10.18653/v1/N18-2026} {Determining event
  durations: Models and error analysis}.
\newblock In \emph{Proceedings of the 2018 Conference of the North {A}merican
  Chapter of the Association for Computational Linguistics: Human Language
  Technologies, Volume 2 (Short Papers)}, pages 164--168, New Orleans,
  Louisiana. Association for Computational Linguistics.

\bibitem[{Vendler(1957)}]{vendler1957verbs}
Zeno Vendler. 1957.
\newblock Verbs and times.
\newblock \emph{The philosophical review}, 66(2):143--160.

\bibitem[{Verkuyl(1972)}]{verkuyl1972compositional}
Henk~J Verkuyl. 1972.
\newblock \emph{On the Compositional Nature of the Aspects.}, volume 15 D of
  \emph{Foundations of Language, Supplementary Series}.
\newblock D. Reidel Publishing Company, Dordrecht-Holland.

\bibitem[{Verkuyl(2005)}]{verkuyl2005aspectual}
Henk~J Verkuyl. 2005.
\newblock Aspectual composition: Surveying the ingredients.
\newblock In \emph{Perspectives on aspect}, pages 19--39. Springer.

\bibitem[{Whorf(1945)}]{whorf1945grammatical}
Benjamin~Lee Whorf. 1945.
\newblock Grammatical categories.
\newblock \emph{Language}, 21(1):1--11.

\bibitem[{Williams(2012)}]{williams2012srw}
Jennifer Williams. 2012.
\newblock {Extracting Fine-grained Durations for Verbs from Twitter}.
\newblock In \emph{Proceedings of the ACL 2012 Student Research Workshop},
  pages 49--54, Jeju Island, Korea.

\bibitem[{Williams and Katz(2012)}]{williams-katz-2012-extracting}
Jennifer Williams and Graham Katz. 2012.
\newblock \href {https://aclanthology.org/P12-2044} {Extracting and modeling
  durations for habits and events from {T}witter}.
\newblock In \emph{Proceedings of the 50th Annual Meeting of the Association
  for Computational Linguistics (Volume 2: Short Papers)}, pages 223--227, Jeju
  Island, Korea. Association for Computational Linguistics.

\bibitem[{Yang et~al.(2019)Yang, Dai, Yang, Carbonell, Salakhutdinov, and
  Le}]{yang2019xlnet}
Zhilin Yang, Zihang Dai, Yiming Yang, Jaime Carbonell, Russ~R Salakhutdinov,
  and Quoc~V Le. 2019.
\newblock Xlnet: Generalized autoregressive pretraining for language
  understanding.
\newblock \emph{Advances in neural information processing systems}, 32.

\bibitem[{Yang et~al.(2020)Yang, Du, Rush, and
  Cardie}]{yang-etal-2020-improving}
Zonglin Yang, Xinya Du, Alexander Rush, and Claire Cardie. 2020.
\newblock \href {https://doi.org/10.18653/v1/2020.findings-emnlp.302}
  {Improving event duration prediction via time-aware pre-training}.
\newblock In \emph{Findings of the Association for Computational Linguistics:
  EMNLP 2020}, pages 3370--3378, Online. Association for Computational
  Linguistics.

\bibitem[{{\v{Z}}abokrtsk{\`y} and
  Lopatkov{\'a}(2007)}]{vzabokrtsky2007valency}
Zden{\v{e}}k {\v{Z}}abokrtsk{\`y} and Mark{\'e}ta Lopatkov{\'a}. 2007.
\newblock Valency information in vallex 2.0.
\newblock \emph{The Prague Bulletin of Mathematical Linguistics}, 87:41--60.

\bibitem[{Zarcone and Lenci(2008)}]{zarcone-lenci-2008-computational}
Alessandra Zarcone and Alessandro Lenci. 2008.
\newblock \href
  {http://www.lrec-conf.org/proceedings/lrec2008/pdf/315_paper.pdf}
  {Computational models for event type classification in context}.
\newblock In \emph{Proceedings of the Sixth International Conference on
  Language Resources and Evaluation ({LREC}'08)}, Marrakech, Morocco. European
  Language Resources Association (ELRA).

\bibitem[{Zellers and Choi(2017)}]{zellers-choi-2017-zero}
Rowan Zellers and Yejin Choi. 2017.
\newblock \href {https://doi.org/10.18653/v1/D17-1099} {Zero-shot activity
  recognition with verb attribute induction}.
\newblock In \emph{Proceedings of the 2017 Conference on Empirical Methods in
  Natural Language Processing}, pages 946--958, Copenhagen, Denmark.
  Association for Computational Linguistics.

\bibitem[{Zhang and Xue(2014)}]{zhang-xue-2014-automatic}
Yuchen Zhang and Nianwen Xue. 2014.
\newblock \href {https://doi.org/10.3115/v1/D14-1204} {Automatic inference of
  the tense of {C}hinese events using implicit linguistic information}.
\newblock In \emph{Proceedings of the 2014 Conference on Empirical Methods in
  Natural Language Processing ({EMNLP})}, pages 1902--1911, Doha, Qatar.
  Association for Computational Linguistics.

\bibitem[{Zhou et~al.(2019)Zhou, Khashabi, Ning, and
  Roth}]{zhou-etal-2019-going}
Ben Zhou, Daniel Khashabi, Qiang Ning, and Dan Roth. 2019.
\newblock \href {https://doi.org/10.18653/v1/D19-1332} {{``}going on a
  vacation{''} takes longer than {``}going for a walk{''}: A study of temporal
  commonsense understanding}.
\newblock In \emph{Proceedings of the 2019 Conference on Empirical Methods in
  Natural Language Processing and the 9th International Joint Conference on
  Natural Language Processing (EMNLP-IJCNLP)}, pages 3363--3369, Hong Kong,
  China. Association for Computational Linguistics.

\end{thebibliography}

\clearpage
\appendix
\section*{APPENDIX}
\label{sec:appendix}

\section{Glossary}\label{sec:glossary}
This glossary intends to provide a concise alphabetically ordered overview of the linguistic terminology used in this paper.

\begin{description}
    \item[accomplishment] eventive / dynamic situation type according to \citep{vendler1957verbs,smith1997parameter}, consisting of a preparatory phase and an endpoint, e.g., \enquote{build a house.}
    \item[achievement]  eventive / dynamic situation type according to \citep{vendler1957verbs,smith1997parameter} consisting only of the \textit{punctual} event that changes a state, e.g., \enquote{win the race.}
    \item[activity] eventive / dynamic situation type according to \citep{vendler1957verbs,smith1997parameter}, describes a process that does not have a pre-defined endpoint, e.g., \enquote{swim.}
    \item[atelic] describes situations or verbs that do not have a built-in endpoint, e.g., \enquote{draw.}
    \item[bounded] see \textit{telic}.
    \item[covert] describes linguistic categories that are not directly inferrable from the lexicon or from grammatical/syntactic structure \citep{whorf1945grammatical}. They usually consist of \textit{semantic} concepts that require taking into account the context or pragmatic factors.
    \item[durative] describes states that extend in time.
    \item[dynamic] describes a type of situation / verb that has a part to its meaning that applies a force or changes a state.%
    \item[episodic] describes particular events that \enquote{happen} or \enquote{have happened} (opposite of \textit{habituals}).
    \item[eventive] see \textit{dynamic}.
    \item[habitual] clause that expresses a situation that repeats regularly and expresses a characteristic, e.g., \enquote{John drives to work} or \enquote{Bishops move diagonally} \citep{krifka1995genericity}.
    \item[imperfective] describes a perspective on a situation focusing one or none of the potential endpoints, but not both 
    \item[iteratives] clauses (usually with \textit{semelfactive} verbs) that signal that the event expressed by the verb is executed a number of times, e.g., \enquote{the light flashed.}
    \item[overt] describes linguistic categories that are directly inferrable from the lexicon or from grammatical/syntactic structures.
    \item[state] a situation that is not changing, e.g., \enquote{John owns a house.}
    \item[stative] describes verbs or situations that express a state.
    \item[perfective] describes constructions that show a situation with its endpoint(s), e.g., \enquote{John traveled to the US (last week).}
     \item[progressive] describes constructions that focus on a particular phase of a situation that is currently ongoing, e.g., \enquote{John is traveling to the US (at the moment).}
    \item[punctual] event that occurs at a single point in time, e.g., \enquote{hiccup.} -- \textit{dynamic} by definition.
    \item[semelfactive] \textit{punctual} situation or event type \citep{smith1997parameter}.
    \item[telic] describes situations or verbs that have a built-in endpoint, e.g., \enquote{capture.}
\end{description}

\section{Overview of Modeling Approaches}
\label{sec:overview-modeling}

\tref{aspect-modeling} gives a concise overview of existing modeling approaches.

\begin{table*}[h]
    \centering
    \footnotesize
    \setlength\tabcolsep{6pt}
    \begin{tabular}{lll}
    \toprule
    \textbf{Type of Model} & \textbf{Reference(s)} & \textbf{Categories Targeted / Task}\\
    \midrule
       Rule-based  & \citet{passonneau1988computational} (PUNDIT) & stativity \\
                    & \citet{chen-etal-2021-autoaspect} & UMR aspect features\\
       Progressive as indicator  & \citet{klavans1992degrees} & stativity \\ 
       Linguistic indicators & \citet{brent1991automatic} & stativity\\
                            & \citet{siegel1996gathering,siegel-mckeown-2000-learning} & stativity, boundedness \\
    Naive Bayes + Decision Tree & \citet{mathew2009supervised} & habituality \\
    Random Forest Classifiers & \citet{friedrich-pinkal-2015-automatic} & habituality \\
    Logistic regression & \citet{friedrich-palmer-2014-automatic} & lexical aspect\\
    & \citet{loaiciga-grisot-2016-predicting} & boundedness\\
    &  \citet{friedrich-gateva-2017-classification} & telicity \\
    & \citet{zarcone-lenci-2008-computational} & Vendler types\\
    
    Distributional models & \citet{kober-etal-2020-aspectuality} & telicity\\
    SVM & \citet{richard2012lexical} & Leech's classes\\
    & \citet{hermes2015automatic,hermes2018supervised} & Vendler classes\\
    Max Ent Sequence Tagger & \citet{palmer-etal-2007-sequencing} & situation entity types\\
    CRF & \citet{friedrich-pinkal-2015-discourse} & genericity\\
    & \citet{friedrich-etal-2016-situation} & situation entities\\
    word2vec + GRU & \citet{becker-etal-2017-classifying} & situation entity types\\
    GloVe + GRU & \citet{zellers-choi-2017-zero} & verb-level Vendler classes\\
    Paragraph-level neural model & \citet{dai-huang-2018-building} & situation entity types\\
    ELMO + MLP & \citet{govindarajan-etal-2019-decomposing} & habituality (continuous features, UDS-G)\\
    BERT + classifiers & \citet{metheniti-etal-2021-predire,metheniti-etal-2022-about} & telicity\\
    BERT + tagger & e.g., \citet{kondratyuk-straka-2019-75} & UD aspectual features\\
    BERT, GPT, ParBERT & \citet{rezaee-etal-2021-discriminative} & situation entity types\\
    Generative model & \citet{gantt2022decomposing} & event types\\
    \bottomrule
    \end{tabular}
     \caption{\textbf{Computational systems and approaches} to modeling aspect.}
    \label{tab:aspect-modeling}
\end{table*}

\section{Further Linguistic Background}

\subsection{Eventuality Types}
\label{sec:eventtypes}

In \sref{vendler}, we explain a taxonomy of eventuality types according to \citet{vendler1957verbs}.
As shown in \fref{aspect:vendler}, \citet{kenny1963action} adds \textit{performances}, which are \enquote{actions that tend towards a goal.}
\citet{mourelatos1978events} criticizes that these earlier analyses focus too much on predicates that require human agency, and suggests the terminology in \fref{aspect:vendler}.

\begin{figure}[h]
\centering
\footnotesize
\begin{tikzpicture}[scale=0.9]
\tikzset{every tree node/.style={anchor=north,align=center}}
\tikzset{edge from parent/.style=
{draw,
edge from parent path={(\tikzparentnode.south)
-- +(0,-4pt)
-| (\tikzchildnode)}}}
\Tree [.situations  states [.occurrences\\(actions) processes\\(activities) [.events\\(performances) developments\\(accomplishments) punctual\ occurrences\\(achievements) ] ] ]
\end{tikzpicture}
\caption{Classification of aspectual oppositions according to \citet{mourelatos1978events}, terms used by \citet{vendler1957verbs} and \citet{kenny1963action} in parentheses.}
\label{fig:aspect:vendler}
\end{figure}
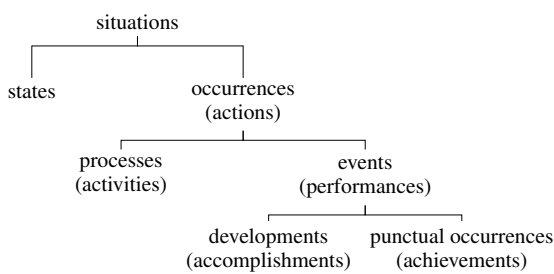

As explained in \sref{vendler}, \citet{moens-steedman-1988-temporal} introduce an inventory of eventuality types similar to those of \citet{vendler1957verbs}, but referring to the characteristics of whether an event is \textit{atomic}, \textit{extended} (\textit{durative}) and whether it has a \textit{consequent state}.
The definitions of the types along with examples are given in \tref{vendlerMoensClasses}.

\begin{table*}[h!]
\centering
\footnotesize
\setlength\tabcolsep{5pt}
\begin{tabular}{lccc|lccc|l}
\toprule
\citet{vendler1957verbs} & \rotatebox[origin=c]{90}{\parbox[c]{1cm}{\textit{dynamic}}} & \rotatebox[origin=c]{90}{\parbox[c]{1cm}{\textit{durative}}} & \rotatebox[origin=c]{90}{\parbox[c]{1cm}{\textit{telic}}} & \begin{tabular}{c}\citeauthor{moens-steedman-1988-temporal}\\ (\citeyear{moens-steedman-1988-temporal}) \end{tabular} & \rotatebox[origin=c]{90}{\parbox[c]{13mm}{\itshape atomic event}} & \rotatebox[origin=c]{90}{\parbox[c]{13mm}{\itshape extended event}} & \rotatebox[origin=c]{90}{\parbox[c]{13mm}{\itshape \centering consequent state}} & \textbf{Examples}\\
\midrule
State & - & + & - & state & NA & NA & NA & know the answer, love Mary, understand\\
Activity & + & + & - & process & - & + & - & laugh, stroll in the park, run, swim\\
Accomplishment & + & + & + & culminated process & - & + & + & build a house, walk to school\\
Achievement & + & - & + & culmination & + & - & + & win a race, reach the top, recognize\\
Semelfactive & + & - & - & point event & + & - & - & tap, knock, hiccup, wink\\
\bottomrule
\end{tabular}
\caption{\textbf{Eventuality} types \citep{vendler1957verbs,smith1997parameter} / \textbf{Temporal Ontology} \citep{moens-steedman-1988-temporal}.}
\label{tab:vendlerMoensClasses}
\end{table*}

\subsection{Leech's Classes}
\label{sec:leech}
In addition to categories corresponding to those of \fref{aspect:vendler}, \citet{leech1971meaning} distinguishes between \textit{Attitude} states (\enquote{hate,} \enquote{hope}) and \textit{Relationship} states (\enquote{own,} \enquote{resemble}).
There are also two categories for \textit{Perception} and \textit{Cognition} verbs such as \enquote{hear,} \enquote{see,} or \enquote{feel,} which are generally hard to classify along the stative-dynamic dimension \citep{comrie1976aspect}.

\subsection{Details on Viewpoint Aspect}
\label{sec:viewpoint}
In some cases, viewpoint aspect (\sref{grammAspect}) and situation type interact.
The perfective viewpoint is naturally available for the situation types (\sref{vendler}) activity, accomplishment, semelfactive and achievement; the imperfective viewpoint is available for states, activities and accomplishments \citep{smith1997parameter}.
The usage of the perfective viewpoint with stative predicates indicates aspectual coercion as introduced
in \sref{vendler}.
For example, in Chinese the stative predicate \enquote{\gao1} means to be tall.
When used with the perfective marker \enquote{le,} an ingressive meaning is implied, i.e., \enquote{\ta1 \gao1-le} translates as \enquote{he became tall.}

\subsection{Habituals: Borderline cases}
\label{sec:habituals-special-cases}

Habitual sentences (\sref{habituality}) describe situations that are characteristic of an extended period of time \citep{comrie1976aspect}, a decision that is of conceptual rather than of linguistic nature.
In fact, \citet{filip1997sui} even argue that \textit{sentential genericity}, which corresponds to habituality, is independent from tense and aspect, and that habitual sentences such as \eref{habitual-perfective} can occur in the perfective. %

\begin{example}
In the eighties, John went to work by bus. %
\label{ex:habitual-perfective}
\end{example}

\noindent
The interaction of habituality and modality is by no means trivial \citep{hacquard2009interaction} as illustrated by \eeref{habitual-unclear}{a}, and negated sentences are another unclear case \eeref{habitual-unclear}{b-c}.

\begin{example}
(a) I had to swim every day. \textit{(habitual?)}\\
\hspace*{10mm} (b) John smokes. \textit{(habitual)}\\
\hspace*{10mm} (c) John does not smoke. \textit{(habitual?)}
\label{ex:habitual-unclear}
\end{example}

\noindent
The concept of habituality does not include \textit{dispositional sentences} such as \enquote{John can swim,} which denote abilities or preferences.

\section{Further Computational Background}
\subsection{Linguistic Indicators}
\label{sec:ling_ind}

\tref{linguistic-indicators} reports the full set of linguistic indicator features as proposed by \citet{siegel-mckeown-2000-learning} and related works.

\begin{table}[h!]
    \centering
    \footnotesize
    \begin{tabular}{ll}
    \toprule
         \textbf{Feature} & \textbf{Example} \\
         \midrule
         frequency & - \\
         past & \textit{said}\\
         perfect & \textit{had won}\\
         progressive & \textit{is winning}\\
         negated & \textit{not/never}\\
         particle & \textit{up/in/...}\\
         no subject & - \\
         continuous adverb & \textit{continually, endlessly}\\
         evaluation adverb & \textit{better, horribly}\\
         manner adverb & \textit{furiously, patiently}\\
         temporal adverb & \textit{again, finally}\\
         in-PP & \textit{in an hour}\\
         for-PP & \textit{for an hour}\\
         \bottomrule
    \end{tabular}
    \caption{\textbf{Linguistic indicators} computed over large syntactically parsed text corpora \citep{siegel-mckeown-2000-learning}.}
    \label{tab:linguistic-indicators}
\end{table}

\subsection{Aspect in UMR} \label{sec:umrAspect}

\begin{figure*}[h]
  \includegraphics[width=1.0\textwidth]{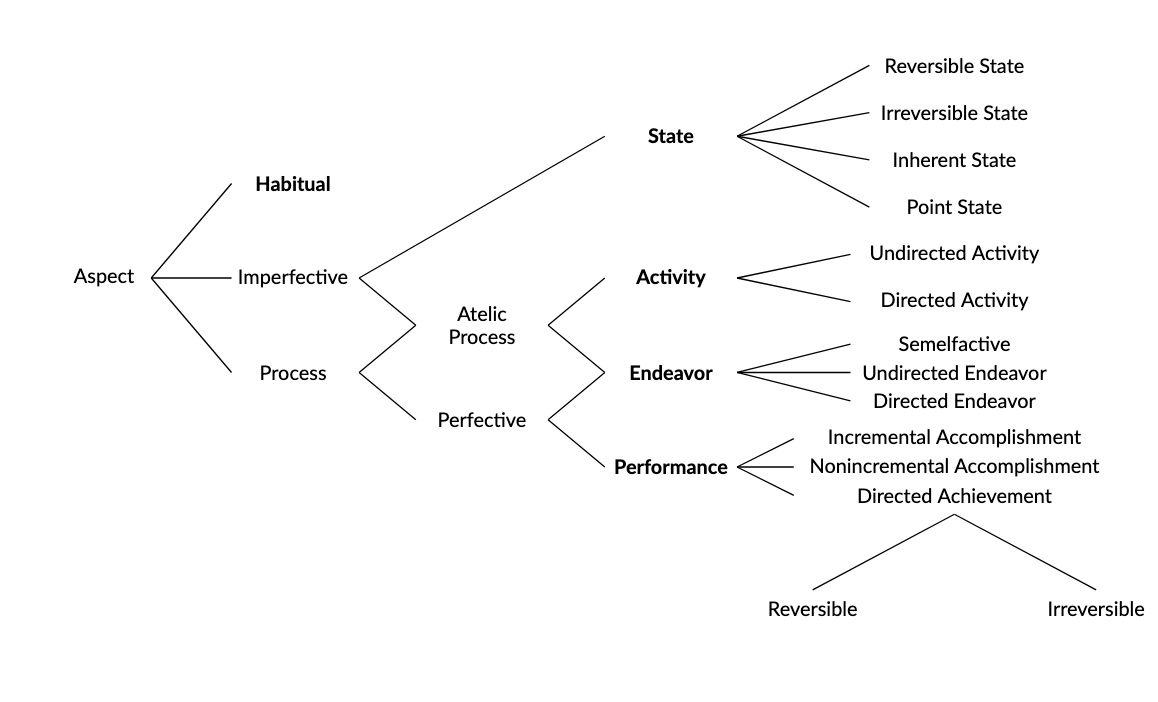}
    \caption{UMR annotation lattice for aspect \citep{vangysel2022}.}\label{umr-aspect-lattice}
\end{figure*}

As briefly described in \sref{grammAspect-frameworks},
the UMR (Uniform Meaning Representation) %
\cite{vanGysel2021}
framework approaches annotation of aspect from a typological perspective. Aspect is represented as an attribute for events in UMR, and since cross-linguistically languages mark aspect in different ways, both grammatically and lexically, the UMR aspect categories are arranged in a lattice. 
This allows UMRs to be annotated at the level of granularity that is most appropriate for a particular language.

The resulting lattice \cite{vangysel2022} appears in Figure \ref{umr-aspect-lattice}. The top-level of the lattice represents broad distinctions between aspect categories across languages, while the bottom level represents the finer distinctions.

\subsection{Situation entity inventory}\label{sec:ses-full}

Table~\ref{tab:se-types-full} provides the complete inventory of situation entity types.
\sref{sitent} offers a partial inventory, focusing on those categories for which aspect is a key determining property.

\begin{table}[t]
    \centering
    \footnotesize
    \begin{tabular}{rl}
    \toprule
    \textbf{SE type} & \textbf{Example}\\
    \midrule
    \multicolumn{2}{l}{\hspace{1cm}\textbf{Eventualities}}\\
        \textsc{State} & The colonel owns the farm. \\
        \textsc{Event} & John won the race.\\
        \textsc{Report} & \enquote{...}, said Obama.\\
        \midrule
        \multicolumn{2}{l}{\hspace{1cm}\textbf{General Statives}}\\
        \textsc{Generic Sent.} & The lion has a bushy tail.\\
        \textsc{Generalizing} & Mary often fed the cat last year.\\
        \hspace{5mm}\textsc{Sentence} & \\
        \midrule
        \multicolumn{2}{l}{\hspace{1cm}\textbf{Abstract Entities}}\\
        \textsc{Fact} & (I know) that she refused the offer.\\
        \textsc{Proposition} & (I believe) that she refused the offer.\\
        \midrule
        \textsc{Question} & Who wants to come?\\
        \textsc{Imperative} & Hand me the pen!\\
        \bottomrule
    \end{tabular}
    \caption{Situation entity types \citep{smith2003modes}. }
    \label{tab:se-types-full}
\end{table}

\subsection{Event Classes in TimeBank}
\label{sec:timeml}

TimeML \citep{pustejovsky2003timeml,pustejovsky-etal-2010-iso} events are \enquote{situations that occur or happen,} but also include \enquote{states or circumstances in which something obtains or holds true.}
Thus, the usage of the term \textit{event} denotes a \textit{situation} in the sense of \citet{smith1997parameter}.
In TimeML, tensed verbs, stative adjectives and event nominals that describe situations temporally located in the text can be marked with the categories shown in \tref{timeml-event-classes}.

TimeML event classification corresponds to a word-sense level task.
\citet{sauri-etal-2005-evita}, in their event recognition system, simply assign the class that was most frequently observed for each verb type in the training data to events and reach an accuracy of 82.3\% on TimeBank 1.2.
The top-performing systems \citep{jung-stent-2013-att1,bethard-2013-cleartk,chambers-2013-navytime} of the 2013 TempEval challenge \citep{uzzaman2013semeval} use corpus-based features, WordNet synsets, parse paths and features from typed dependencies to classify events as a joint task with determining the event’s span.
\citet{bethard-martin-2006-identification} phrase the recognition of EVENTs and their semantic class as a chunking task using syntactic-semantic features such as part-of-speech, morphological information, word clusters and WordNet hypernyms.
\citet{llorens-etal-2010-timeml} extend this idea by using a conditional random field \citep{lafferty2001conditional} enhanced with semantic role information.
\citet{costa-branco-2012-aspectual} explore the usefulness of a wider range of explicitly aspectual features, including linguistic indicators, for temporal relation classification in Portuguese.

More recent TempEval challenges \citep{bethard-etal-2016-semeval,bethard-etal-2017-semeval} did not offer a task for classifying event classes.

\begin{table}[t]
    \centering
    \footnotesize
    \begin{tabular}{rl}
    \toprule
    \textbf{Event Class} & \textbf{Explanation and Examples}\\
    \midrule
        OCCURRENCE & situations that happen: \\
        & \textit{die, crash, merge}\\
        STATE & circumstances in which sth. holds: \\
        & \textit{like, own, the kidnapped girl, on board}\\
        I\_ACTION & intensional actions: \textit{try, persuade}\\
        I\_STATE & intensional states: \textit{love, believe, enjoy}\\
        ASPECTUAL & predicates that pick out a phase\\
        & of the event: \textit{begin, start, continue}\\
        I\_REPORTING & capture attribution: \textit{said}\\
        PERECPTION & physical perception of another event:\\
        & \textit{see, hear, feel}\\
        \bottomrule
    \end{tabular}
    \caption{TimeML event classes \citep{pustejovsky2003timeml}.}
    \label{tab:timeml-event-classes}
\end{table}

\section{(Potential) Applications}
\label{sec:potential_applications}

Despite its importance for understanding, few NLP tasks explicitly incorporate aspectual information. 
Here we discuss the potential uses in two application areas; many other areas could also benefit, including argumentation mining, computer-aided language learning, and information extraction.

\paragraph{Temporal Processing.}
Together with tense, aspect is essential to the linguistic system encoding temporal information \citep{smith1997parameter,smith2003modes}.
Yet there is little work systematically leveraging aspectual information in temporal relation extraction systems, possibly 
because TimeML event classes \citep[][\aref{timeml}]{pustejovsky2003timeml,pustejovsky-etal-2010-iso} are not strictly rooted in linguistic theory.
We believe that recognizing the mode of discourse and the type of temporal progression (which is different e.g., in a narrative, an encyclopedia entry, or a news article) is key to 
recognizing temporal structure in text.

With the help of templates, \citet{vashishtha-etal-2020-temporal} re-cast existing temporal relation and temporal duration datasets as natural language inference (NLI) datasets.
Existing models perform well on classifying English sentence-internal relations this way, indicating that the number of tense pair patterns is somewhat limited. %
However, using large-pretrained transformer based models with a paradigm shift \citep{sun2021paradigmshift} is unlikely to be the solution to temporal reasoning, as shown by \citet{kober-etal-2021-data}.
They create a dataset consisting of 11,138 pairs of short sentences labeled with a binary annotation scheme (\textit{entailment} vs. \textit{non-entailment}), addressing tense and the perfect vs. progressive distinction in English.
For example, \enquote{John is visiting London} entails \enquote{John has arrived in London,} but \enquote{John will visit London} does not.
They evaluate a range of models including non-contextualized and contextualized embeddings pre-trained on SNLI \citep{bowman-etal-2015-large} and DNC \citep{poliak-etal-2018-collecting} and find that none of these models outperforms a majority class baseline on the new dataset.
This suggests that embedding spaces struggle to capture the more latent nature of tense and aspect.

In our view, a first step towards leveraging the knowledge about tense and aspect provided by linguistic theory could be to systematically study how recent language models and NLP systems succeed or fail with regard to these categories, as 
pioneered in the pre-neural age for example by \citet{zhang-xue-2014-automatic,reichart-rappoport-2010-tense}.

\paragraph{Machine Translation.}

While recent MT systems often perform well, they do not systematically treat aspectual notions \citep{vanmassenhove2017investigating}, but rather rely on common translations for a particular domain. %
Consider the following example.
In contrast to English, the German simple present may or may not indicate habituality. %
When translating to English, a choice must be made based on the context.
The translation\footnote{Google Translate, October 5, 2022} in \eeref{mt-example2}{b} wrongly indicates habitual viewpoint, while the context suggests an ongoing event (the correct translation would be \enquote{is riding}).

\begin{example}
(a) Tim ist schon weg. Er \dul{fährt} gerade mit\\\hspace*{14mm} dem Fahrrad zur Arbeit. \textit{(ongoing event)}\\
\hspace*{9mm}(b) Tim is already gone. He just \dul{rides} his\\ \hspace*{14mm} bike to work. \textit{(habitual, wrong)}
\label{ex:mt-example2}
\end{example}

\noindent
The cue need not be in the immediately preceding sentence, but could occur much earlier in the discourse.
In \eref{mt-example2long}, \enquote{fährt} should be translated as \enquote{is riding,} but the cue is in the first sentence.

\begin{example}
(a) Tim ist schon weg. Er arbeitet an der\\ \hspace*{14mm} Universität. Er \dul{fährt} gerade mit dem\\\hspace*{14mm} Fahrrad zur Arbeit. \textit{(ongoing event)}\\
\hspace*{9mm}(b) Tim is already gone. He works at the\\ \hspace*{14mm} university. He just \dul{rides} his bike to work. \\ \hspace*{14mm} \textit{(habitual, wrong\footnote{Google Translate, October 19, 2022})}
\label{ex:mt-example2long}
\end{example}

\noindent
Aspect is a crucial component of semantic interpretation, and special care is needed to make sure translations correctly capture it.

\section{ChatGPT Logs}
\label{sec:chat-gpt-logs}

Accessed January 26 and 27, 2023.

In this section we provide logs from our interactions with ChatGPT on some of the main topics covered by this survey. 
Our aim is to investigate the extent to which a system like ChatGPT could serve as a tool for teaching about aspect; in other words, what does ChatGPT know about aspect?

Note that some of ChatGPT's analyses are surprisingly accurate, however, it also presents several completely wrong analyses and only produces correct analyses if guided to them by the user during the generation process.

Each subsection contains one or more segments from a chat log about a particular aspectual notion. 
We use the formatting shown below:
\begin{itemize}
    \item \textit{Prompt}
    \item Text generated by ChatGPT (with interesting parts highlighted in \textbf{bold})
    \item \textit{\comm{Comments}}
\end{itemize}

\subsection{Talking about Stativity}

\textit{Consider the following text. "Alice was beginning to get very tired of sitting by her sister on the bank, and of having nothing to do: once or twice she had peeped into the book her sister was reading, but it had no pictures or conversations in it, and `what is the use of a book,' thought Alice, `without pictures or conversation?'" Which verb phrases refer to states, which verb phrases refer to events in this passage?}

In this passage, "was beginning to get very tired" and "having nothing to do" are verb phrases that refer to states. "She had peeped" and \textbf{"had no pictures or conversations in it" are verb phrases that refer to events}.

\comm{The verb phrases "have nothing to do" and "had no pictures or conversations in it" are states; "had peeped" is an event; "was beginning" is a special case at it signals the beginning of a state (\textit{inchoative} aspect). ChatGPT's response is simplistic and in some cases completely wrong.}

\textit{Why is "had no pictures or conversations in it" an event?}

\textbf{The phrase "had no pictures or conversations in it" is an event because it describes something that happened in the past}, specifically that the book did not contain pictures or conversations. The verb "had" is in the past tense which indicates that \textbf{the action of the book not having pictures or conversations} is completed, and it happened in the past. An event can be a completed past action or something that is happening currently or in the future.

[...]

\textit{Could you interpret the phrase as a state as well?}

It could be interpreted as a state in the sense that the book currently lacks pictures or conversations, however, in the context of the passage \textbf{it is clear that the author is describing a past event} where Alice had looked into the book and found that it did not contain pictures or conversations, \textbf{as opposed to the book being in a state of lacking pictures or conversations}. The past tense of the verb "had" and the context of Alice peeping into the book suggest it's an event that happened in the past. Additionally, the phrase "`what is the use of a book,' thought Alice, `without pictures or conversation?'" also suggest that it's an event that happened in the past. So, it can be interpreted as a state, but in this context, it is more likely an event.

\comm{Even in this context, the verb phrase describes a state (a property of the book).}

\textit{I agree that "peeped into the book" is an event. But what about "the book had no pictures in it"?}

\textbf{In the sentence "the book had no pictures in it", the phrase "had no pictures in it" describes the state of the book, specifically that it did not contain any pictures.} The verb "had" is in the past tense, which indicates that this is a statement about the book's state at a specific point in the past. The phrase describes a state that existed at a certain point in the past. Even though "had" is in past tense, it is used to describe a state, not an event. Additionally, the phrase is not describing something that just happened, but a characteristic of the book, which implies a state.

\comm{This explanation is correct. However, the above conversation clearly illustrates that ChatGPT generates text explaining the aspectual nature of the various verb phrases without having a deeper understanding of their semantics and hence without being consistent in its relatively random interpretations.}

\textit{Telicity is a property of an event type expressed by verbs, i.e., referring to events that have an endpoint. Atelic verbs refer to event types that do not have a built-in endpoint. Can you give me examples for each case?}

Sure, here are some examples of telic verbs and the events they describe:

"She finished the book." (finish)

"He graduated from college." (graduate)

"They completed the project." (complete)

And here are some examples of atelic verbs and the events they describe:

"She is reading a book." (read)

"He is studying for the exam." (study)

"They are working on the project." (work)

\comm{These examples are correctly classified except for "read a book."}

\textit{Why is the event type "read a book" atelic?}

\textbf{The event type "read a book" is considered atelic because it does not have a built-in endpoint.} In other words, when someone says "She is reading a book," it does not necessarily imply that the act of reading will come to a definite conclusion. Reading a book can be an ongoing activity that may or may not be completed. \textbf{It does not have an inherent endpoint.}

\comm{Here, the analysis confuses between progressive aspect and analysing telicity at the event type (predicate) level. (Admittedly, this is a very fine-grained notion that humans also need to think carefully about.) In "She was reading a book", the event type "read a book" is telic, while the sentence in the progressive presents the ongoing situation of reading a book, hiding the end points.}

\textit{Wouldn't the endpoint mean she finished reading the entire book?}

Yes, in the sentence "She finished the book" the endpoint would mean she finished reading the entire book. The verb "finish" implies that an action has come to completion, so when it is used in the context of reading a book, it implies that the book has been read in its entirety. However, it's worth noting that some verbs can have multiple meanings and context can affect the interpretation of a sentence.

In contrast, when someone says "She is reading a book" it doesn't necessarily imply that the act of reading will come to a definite conclusion or that she will finish reading the entire book, she could stop reading at any time. And that's why the event is atelic.

\textit{I think the event is still telic, but the sentence focuses on a phase of it.}

You are correct that the event of reading a book, in general, is telic, as it does have a natural endpoint, which is when the person finishes reading the book. However, when we use the verb "read" in the present progressive form "is reading," it describes the action as being ongoing and does not imply a definite conclusion. It describes the event as a phase or a state of being rather than a completed action.

So in the sentence "She is reading a book," the emphasis is on the current state of reading and not on the endpoint of finishing the book. The sentence "She finished the book" implies that the action of reading is completed and the verb "finish" implies that an action has come to completion.

\textbf{So, in summary, while the overall event of reading a book is telic, in certain sentence constructions, such as "She is reading a book," the focus is on the ongoing state of the event, making it atelic.}

\comm{The last part of the analysis is correct, except that the event of reading a book is always telic, and the progressive focuses on part of the situation. This does not make the event type atelic. The explanations given in this section focus more on the progressive vs. non-progressive distinction, not on the telic vs. atelic distinction.}

\subsection{Eventuality Types}

\textit{(Prompted with the first two paragraphs of \sref{vendler} of this paper including examples.)}

\textit{Can you please analyse: "Tom held a meeting"}

The sentence \textbf{"Tom held a meeting" is an example of an achievement} according to Vendler's (1957) classification of situation types. The sentence describes an event that has an endpoint, the meeting, and it is also a change of state. The verb "held" is also telic, meaning it has a specific endpoint and it does not include a process leading up to this point, which fits with the definition of an achievement. The phrase "held a meeting" implies that there was a start and an end to the meeting and it implies that it has been completed.

\textit{Can you please analyse: "Tom was holding a meeting"?}

The sentence \textbf{"Tom was holding a meeting" is an example of an activity} according to Vendler's (1957) classification of situation types. The sentence describes an ongoing process of holding a meeting without any specific endpoint. The verb "was holding" is also atelic, meaning it does not have a specific endpoint, which fits with the definition of an activity. The phrase "holding a meeting" implies that there is a meeting in progress, but it does not specify when or if the meeting is completed.

[...]

\comm{The predicate "hold a meeting" is in fact an accomplishment, as it is dynamic, durative, and telic. The change in grammatical aspect from perfect to progressive does not change the underlying eventuality type. Again, the system seems to be misled by the progressive aspect.}

\textit{Can you please analyse: "Anna was swimming across the lake"?}

The sentence "Anna was swimming across the lake" is an example of an activity according to Vendler's (1957) classification of situation types. The sentence describes an ongoing process of swimming without any specific endpoint. The verb "was swimming" is also atelic, meaning it does not have a specific endpoint, which fits with the definition of an activity. \textbf{The phrase "across the lake" indicates that Anna is swimming over a significant distance, covering a large geographical area,} but the sentence does not specify when or if the journey is completed.

\comm{The default interpretation for this sentence is as an accomplishment, with the goal of reaching the other side of the lake. There is a second plausible, though less likely, interpretation in which Anna is engaged in the activity of swimming in a location that is across the lake from the speaker. ChatGPT proposes a third, highly-unlikely interpretation, akin to describing something as happening "all across the world."}

\textit{Can you please analyse "Tim swam across the pool three times".}

The sentence \textbf{"Tim swam across the pool three times" is an example of an accomplishment} according to Vendler's (1957) classification of situation types. The sentence describes a process (swimming) that leads to a built-in terminal point, the completion of swimming across the pool. The verb "swam" is telic, meaning it has a specific endpoint, which fits with the definition of an accomplishment. Additionally, the phrase "across the pool" indicates that Tim is swimming over a significant distance, covering a large geographical area and the sentence implies that the journey has been completed. The number "three times" indicates that the process was repeated a specific number of times, and it gives the sense that the accomplishment has been reached multiple times.

\comm{When substituting \enquote{pool} for \enquote{lake} and adding the frequency adverb \enquote{three times}, GPT's interpretation of the event type of "swim across X" is correct, again hinting at the randomness of the generated explanations.}

\end{document}